\pdfoutput=1
\documentclass{article}
\usepackage{microtype}
\usepackage{graphicx}
\usepackage{booktabs}
\usepackage{amsmath,amssymb}
\usepackage{amsthm}
\usepackage{bm}

\newcommand{\mat}[1]{\mathbf{#1}}
\newcommand{\sys}{MobileSpec}
\usepackage{xcolor}
\usepackage{colortbl}
\usepackage{tabularx}
\usepackage{multirow}
\usepackage{algorithm}
\usepackage{algorithmic}
\definecolor{oursbg}{HTML}{EBF1F8}
\definecolor{posgreen}{HTML}{1A7F37}
\definecolor{negred}{HTML}{C62828}
\newcommand{\posd}[1]{\textcolor{posgreen}{(#1)}}
\newcommand{\negd}[1]{\textcolor{negred}{(#1)}}
\usepackage{hyperref}
\usepackage{xurl}
\usepackage[accepted]{mlsys2025}
\makeatletter
\renewcommand{\Notice@String}{Preprint.}
\renewcommand{\printAffiliationsAndNotice}[1]{%
\stepcounter{@affiliationcounter}%
{\let\thefootnote\relax\footnotetext{\hspace*{-\footnotesep}#1%
\forloop{@affilnum}{1}{\value{@affilnum} < \value{@affiliationcounter}}{
\textsuperscript{\arabic{@affilnum}}\csname @affilname\the@affilnum\endcsname}.

\ \\
\Notice@String}}}
\makeatother
\hypersetup{pdftitle={Training Specialist LLMs on Your Phone: Forward-Only Optimization via Block-Spectral Coordinates},pdfsubject={}}

\newtheorem{proposition}{Proposition}
\newtheorem{corollary}{Corollary}

\usepackage[normalem]{ulem}

\usepackage{tcolorbox}
\newcounter{observation}
\newcounter{takeaway}
\newtcolorbox{keybox}{colback=oursbg,colframe=black!55,boxrule=0.5pt,arc=1.5pt,left=4pt,right=4pt,top=2pt,bottom=2pt,before skip=4pt,after skip=4pt}
\newcommand{\observation}[1]{\refstepcounter{observation}\begin{keybox}\textbf{Observation~\theobservation:} \emph{#1}\end{keybox}}
\newcommand{\takeaway}[1]{\refstepcounter{takeaway}\begin{keybox}\textbf{Takeaway~\thetakeaway:} \emph{#1}\end{keybox}}

\mlsystitlerunning{Training Specialist LLMs on Your Phone: Forward-Only Optimization via Block-Spectral Coordinates}

\begin{document}
\twocolumn[
\mlsystitle{Training Specialist LLMs on Your Phone: Forward-Only Optimization via Block-Spectral Coordinates}
\mlsyssetsymbol{equal}{*}
\begin{mlsysauthorlist}
\mlsysauthor{Liyan Tan}{equal,ece}
\mlsysauthor{Yequan Zhao}{equal,ece}
\mlsysauthor{Ziyue Liu}{cs}
\mlsysauthor{Yupeng Su}{cs}
\mlsysauthor{Ruijie Zhang}{cs}
\mlsysauthor{Zheng Zhang}{ece}
\end{mlsysauthorlist}
\mlsysaffiliation{ece}{Department of Electrical and Computer Engineering, University of California, Santa Barbara, CA, USA}
\mlsysaffiliation{cs}{Department of Computer Science, University of California, Santa Barbara, CA, USA}
\mlsyskeywords{zeroth-order optimization, post-training, spectral adaptation}
\vskip 0.3in

\begin{abstract}
Large language models (LLMs) on personal devices keep user data local, but limited memory restricts their general capability. Since most user-specific tasks do not require the breadth of frontier models, a practical solution is to adapt a shared on-device model into lightweight specialists. However, backpropagation is prohibitive in memory, and users rarely provide labeled data. We present \textbf{\sys{}}, a system that trains specialists with forward passes alone and learns from automatically checked task outcomes, as in reinforcement learning with verifiable rewards (RLVR). Because forward-only updates are built from random perturbations, training is efficient only when these perturbations resemble the weight changes of RLVR, which largely preserve the singular values of pretrained weights and rotate their singular vectors. The core algorithm of \sys{}, \textbf{FuRA-S}, produces such perturbations while training only the block-spectral coordinates of pretrained weights, namely the singular values of their column blocks. As we show theoretically and empirically, rescaling these blocks independently shifts their relative contributions to each weight matrix and thereby rotates its singular vectors. Since this update can be rewritten as a transformation of the layer input, FuRA-S also adapts INT4 models without modifying their weights. With specialists of only 1--5\,MB that update 0.015--0.05\% of the parameters, \sys{} raises the success rate of a phone-use agent from 6--23\% up to 89--100\% and trains a personalized 8B model on an iPhone 16 Pro overnight. On RLVR benchmarks, FuRA-S is competitive with full-parameter evolution strategies (ES) and outperforms ES over LoRA adapters and quantized ES on INT4 models.
\end{abstract}

]
\printAffiliationsAndNotice{\mlsysEqualContribution}

\section{Introduction}
\label{sec:intro}

Large language models (LLMs) are increasingly deployed as autonomous agents on personal devices, from desktop environments~\cite{xie2024osworld} to mobile interfaces~\cite{wang2024mobileagent}. With the rapid progress of small language models~\cite{liu2024mobilellm} and the dedicated inference accelerators (NPUs) in recent smartphones, running such agents entirely on the device has become practical. These agents keep user data local and work offline.

However, the limited memory of mobile devices restricts on-device models to a few billion parameters, leaving their general capability well below that of frontier models. In practice, users rarely need such breadth, since the tasks they perform on the device are highly user-specific. This motivates a deployment paradigm in which a shared base model, referred to as the \emph{generalist}, is adapted locally into lightweight \emph{specialists}. Each specialist is a small set of parameters trained for one of these tasks.

\begin{figure}[t]
\centering
\vspace{-10pt}
\includegraphics[width=\columnwidth]{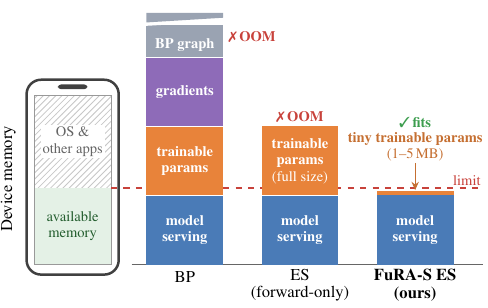}
\vspace{-20pt}
\caption{Memory needed to train on a phone. Serving the model alone occupies most of the memory available to an app, so training with backpropagation or with full-parameter ES exceeds it, whereas FuRA-S adds only 1--5\,MB of trainable parameters.}
\vspace{-5pt}
\label{fig:memory}
\end{figure}

Training specialists on the device, however, imposes constraints that standard post-training pipelines do not satisfy. Since labeled data are scarce, the model should learn from signals that arise during use. These signals include automatic checks of task outcomes, as used in reinforcement learning with verifiable rewards (RLVR;~\citealp{lambert2024tulu3}), and the records that users accumulate on the device. Moreover, mobile NPUs may not afford backpropagation, which consumes several times the memory of inference~\cite{malladi2023mezo} and is poorly supported by the quantized dataflow of on-device engines~\cite{zhao2024poorman}. Forward-only methods, including zeroth-order (ZO) optimization~\cite{malladi2023mezo,zhao2024poorman} and evolution strategies (ES;~\citealp{qiu2025es,sarkar2026eggroll}), avoid these costs by estimating updates from the losses or rewards of randomly perturbed models. Yet they still perturb a large number of model parameters and maintain a full-precision copy of the weights, which exceeds the memory available to an app (Figure~\ref{fig:memory}) and makes every specialist as large as the model itself. This raises a natural question: can a phone still train its own specialists from the signals that arise during use, with forward passes alone?

\begin{figure*}[t]
\centering
\includegraphics[width=\textwidth]{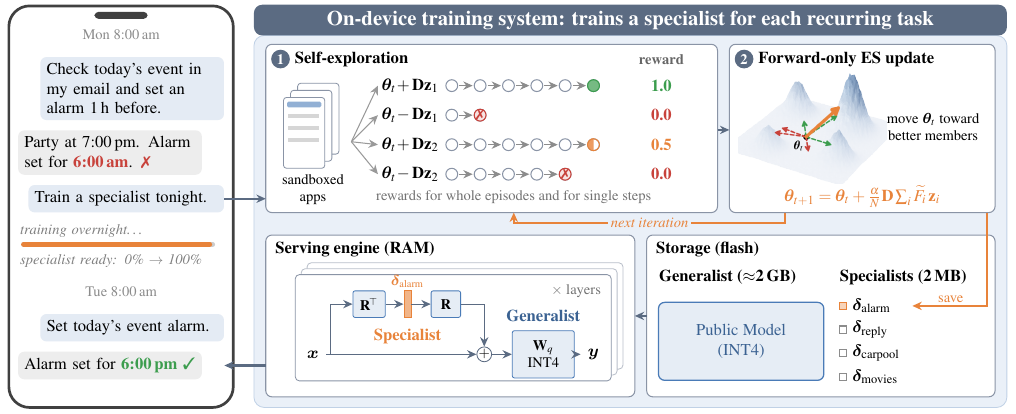}
\caption{On-device training of a specialist with \sys{}. When the generalist agent fails at a task that the user repeats, \sys{} trains a specialist for it on the phone overnight. Perturbed models attempt the task in sandboxed apps, and forward-only ES moves the block singular values toward those of the better perturbed models (Section~\ref{sec:optim}). The resulting specialist is a 2\,MB vector $\bm{\delta}$ that the serving engine applies on top of the shared INT4 generalist.}
\label{fig:overview}
\end{figure*}

In this work, we investigate this question and present \textbf{\sys{}}, an on-device training system that realizes the generalist--specialist paradigm (Figure~\ref{fig:overview}). Its specialists share one INT4 generalist and differ only in files of 1--5\,MB, so switching tasks leaves the generalist and its kernels unchanged. When the phone is idle, \sys{} trains each specialist with forward passes, measuring the fitness of the perturbed models by automatic checks of task outcomes or by the likelihood of the user's records. For long-horizon phone-use tasks, we further design a step-wise fitness function that credits each action by comparing it with the actions that other perturbed models take on the same screen.

How efficiently \sys{} learns depends on which parameters its specialists change. Parameter-efficient fine-tuning (PEFT) methods such as LoRA~\cite{hu2021lora} are a natural choice, but they were designed for backpropagation. Backpropagation computes the exact gradient for any parameterization, whereas forward-only optimization estimates the update as a weighted sum of random perturbations. Learning is therefore efficient only when these perturbations resemble the weight changes that RLVR makes. Recent work characterizes these changes as \emph{spectral inheritance}: RLVR largely preserves the singular values of pretrained weights~\cite{zhu2026iso,zhu2025pathnottaken} and adapts the model mainly by rotating their singular vectors~\cite{zhu2026iso}. The generic low-rank perturbations of LoRA are not designed to produce such rotations. The parameters that existing methods train are thus either too large to store for every task or not aligned with the weight changes of RLVR.

At the core of \sys{} is \textbf{FuRA-S} (\textbf{Fu}ll-\textbf{R}ank \textbf{A}daptation--\textbf{S}pectra), an extremely parameter-efficient method for forward-only training. FuRA-S trains only the block-spectral coordinates of pretrained weights, namely the singular values of their column blocks under an untruncated singular value decomposition (SVD). Yet its ES perturbations change the weights in the way RLVR does. Theoretically, FuRA-S achieves this because rescaling the singular values of each block independently shifts the relative contributions of the blocks to the assembled matrix and thereby rotates its singular vectors rather than changing its singular values. Empirically, what FuRA-S learns comes almost entirely from these rotations: the resulting singular vectors combined with the pretrained singular values retain over 95\% of the accuracy gain. The untruncated factorization also lets FuRA-S adapt the INT4 generalist without modifying its weights, since the update can be rewritten exactly as a transformation of the layer input.

Our contributions are summarized as follows:
\begin{itemize}\setlength{\itemsep}{1pt}
\item We design \sys{}, an on-device training system that realizes the generalist--specialist paradigm. Its specialists of 1--5\,MB share one INT4 generalist, and its step-wise fitness function credits the individual actions of long-horizon phone-use tasks.
\item We propose FuRA-S, the core algorithm of \sys{}, which trains only the block-spectral coordinates of pretrained weights, about 0.03\% of the parameters of Qwen2.5-1.5B, and adapts released INT4 models without modifying their weights.
\item We show through a first-order analysis that FuRA-S changes pretrained weights in the way RLVR does, rotating their singular vectors while largely preserving their singular values, and confirm empirically that the weights it trains exhibit spectral inheritance.
\item We conduct experiments on phone-use, personalization, reward-based, supervised, and agentic tasks. \sys{} raises the success rate of a phone-use agent from 6--23\% up to 89--100\%. On RLVR benchmarks, FuRA-S is competitive with full-parameter ES, consistently outperforms ES over LoRA adapters (LoRA-ES), and surpasses quantized evolution strategies (QES) by up to 16.25 points.
\end{itemize}

\section{Related Work}
\label{sec:related}

\paragraph{Mobile agents and on-device training.}
LLM agents operate phones through their screens, either by locating interface elements with visual perception tools~\cite{wang2024mobileagent} or, as in foundation GUI agents such as MAI-UI~\cite{zhou2025maiui}, by acting directly on screenshots. MobileWorld~\cite{kong2025mobileworld} evaluates them on long tasks that often span several apps. Online reinforcement learning trains them in many parallel environments~\cite{zhou2025maiui,tang2026clawgui}, and GiGPO~\cite{feng2025gigpo} credits individual steps by grouping the actions that different trajectories take from the same state. To fine-tune models on the phone itself, memory-efficient backpropagation reduces the memory of the backward pass~\cite{song2025mebp,park2026mesp} or adapts it to device constraints through sharding and energy-aware scheduling~\cite{geng2025mobilefinetuner}, whereas PocketLLM~\cite{peng2024pocketllm} and MobiZO~\cite{gao2025mobizo} avoid it with forward-only optimization. For personalization, OPPU~\cite{tan2024oppu} trains a separate LoRA adapter for each user.

\paragraph{Forward-only optimization.}
MeZO~\cite{malladi2023mezo} fine-tunes LLMs with the memory footprint of inference, and later ZO methods refine its gradient estimate with low-rank, sparse, or per-example perturbations (LOZO~\cite{chen2024lozo}, Sparse-MeZO~\cite{liu2024sparsemezo}, GRZO~\cite{tan2026grzo}) or lower the precision of its forward passes (QuZO~\cite{zhou2025quzo}). ES at Scale~\cite{qiu2025es} applies full-parameter ES to RLVR of billion-parameter LLMs, and other work extends ES to agentic tasks~\cite{zheng2026agenticesopt}, quantized weights (QES~\cite{xu2026qes}), and the singular values of LoRA adapters~\cite{korotyshova2025essa}, or scales it to large populations with low-rank perturbations (EGGROLL~\cite{sarkar2026eggroll}).

\paragraph{Parameter-efficient fine-tuning.}
LoRA~\cite{hu2021lora} learns low-rank updates, and VeRA~\cite{kopiczko2023vera} and Uni-LoRA~\cite{li2025unilora} reduce its parameters further through shared random projections. Spectral methods initialize LoRA from the principal or minor singular components of pretrained weights~\cite{meng2024pissa,wang2025milora}, tune only the singular values (SVF~\cite{sun2022svf,sun2025transformer2}, SVFit~\cite{sun2024svfit}), or learn sparse combinations of singular vectors~\cite{lingam2024svft}. In RLVR, SVD-informed initializations suffer from spectral collapse, and extreme parameter reduction limits reasoning capacity~\cite{yin2025perlvr}. FuRA~\cite{zhao2026fura} factors each column block of a weight matrix by an untruncated SVD and trains the block singular values together with the right factor of each block, which enables full-rank updates.

\section{The \sys{} System}
\label{sec:system}

\sys{} realizes the generalist--specialist paradigm by keeping one INT4 generalist on the device and training a specialist for each of the user's tasks with forward passes alone while the phone is idle (Figure~\ref{fig:overview}). Its training algorithm, FuRA-S, changes only the block singular values of the generalist and runs directly on its INT4 weights.

\subsection{Workflow}
\label{sec:workflow}

The generalist comprises the INT4 weights and frozen block factors that are computed from them once on a server and distributed with the model (Section~\ref{sec:quant-method}). Each specialist is a 1--5\,MB file of relative changes to the block singular values.

\paragraph{Task and fitness.} Each task needs a fitness function that the device can compute without labels. For a task that ends in a verifiable state, such as replying to a post, the perturbed models attempt it in sandboxed apps, and a checker inspects the final state of the app and returns a reward. For a task defined by the user's own records, such as tagging movies, the fitness of a perturbed model is the likelihood that it assigns to these records.

\paragraph{Training.} When the phone is idle, for example overnight, \sys{} trains the specialist with FuRA-S (Section~\ref{sec:furas}). Each iteration evaluates a population of perturbed models, called members, and moves the specialist toward the members with higher fitness, which for long-horizon tasks is the step-wise fitness of Section~\ref{sec:stepwise}. These forward evaluations stream the decoder layers from flash, so the device never holds the entire model in memory.

\paragraph{Selection.} \sys{} periodically evaluates the specialist on held-out instances of the task, such as the user's later records or new episodes. It retains the best-scoring checkpoint and stops training once the score no longer improves.

\paragraph{Serving.} When the user runs a task, the device loads its specialist and applies it as a transformation of the input before each INT4 multiplication (Section~\ref{sec:quant-method}). Switching to another task therefore replaces only the specialist, while the generalist and its kernels stay unchanged.

\subsection{Training specialists with FuRA-S}
\label{sec:furas}

\begin{figure*}[t]
\centering
\includegraphics[width=\textwidth]{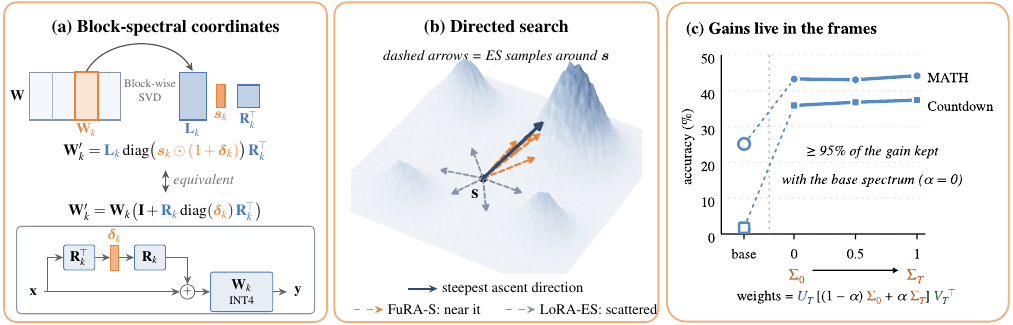}
\caption{FuRA-S. (a)~Only the block singular values are trained, and the adapted block reuses the original, possibly INT4, weights. (b)~FuRA-S restricts its perturbations to the region where more of them improve the model, whereas LoRA perturbs in unconstrained, less effective directions. (c)~The trained singular vectors alone retain over 95\% of the gain.}
\label{fig:algorithm}
\end{figure*}

FuRA-S trains each specialist by ES over the block singular values of the generalist (Figure~\ref{fig:algorithm}a). They make up about 0.03\% of the parameters of Qwen2.5-1.5B and form the search space, which is the set of parameters that ES perturbs. Although only these values change, their perturbations rotate the singular vectors of each weight matrix while barely changing its singular values, as RLVR does. Section~\ref{sec:method} shows that this property lets ES find improving perturbations for longer and learn faster than in other search spaces.

\paragraph{Block-wise SVD reparameterization.}
\label{sec:coords}
Following FuRA \cite{zhao2026fura}, we split $\mat{W}\in\mathbb{R}^{m\times n}$ into $K$ column blocks of width $b$, $\mat{W}=[\,\mat{W}_1\mid\cdots\mid \mat{W}_K\,]$, and factor each block by its untruncated SVD:
\begin{equation}
\mat{W}_k = \mat{L}_k\operatorname{diag}(\mat{s}_k)\mat{R}_k^\top,
\label{eq:block-svd}
\end{equation}
where $\mat{L}_k\in\mathbb{R}^{m\times b}$ has orthonormal columns, $\mat{R}_k\in\mathbb{R}^{b\times b}$ is orthogonal, and $\mat{s}_k\in\mathbb{R}^{b}$ holds the block singular values. Because every block keeps all of its singular values, the factorization reconstructs $\mat{W}$ exactly. The product $\mat{L}_k\operatorname{diag}(\mat{s}_k)$ has the same size as $\mat{W}_k$ and takes its place, so the only overhead in storage and computation comes from the $K$ factors $\mat{R}_k$ of size $b\times b$: $Kb^2=nb$ values per matrix, or a fraction $b/m$ of $\mat{W}$. Partitioning the columns as in FuRA (Appendix~\ref{app:setup}) keeps this overhead at about 3\%.

\paragraph{Update rule.}
\label{sec:optim}

ES updates the parameters from the fitness of perturbed models instead of gradients. At iteration $t$, it evaluates $N$ members $\bm{\theta}_t+\mu\mat{z}_j$ with $\mat{z}_j\sim\mathcal{N}(0,\mat{I}_d)$, obtains their fitness $F_j$, such as a task reward or a negative loss, from forward passes, and updates $\bm{\theta}_{t+1}=\bm{\theta}_t+\frac{\alpha}{N}\sum_{j}F_j\mat{z}_j$, where the step size $\alpha$ absorbs $1/\mu$. 

FuRA-S changes the search space and how the perturbations are sampled and weighted. FuRA-S freezes $\mat{L}_k$ and $\mat{R}_k$ and confines the trainable parameter set $\bm{\theta}$ to the block singular values. The adapted matrix is
\begin{equation}
\mat{W}'=\bigl[\,\mat{L}_1\operatorname{diag}(\mat{s}'_1)\mat{R}_1^\top \;\big|\; \cdots \;\big|\; \mat{L}_K\operatorname{diag}(\mat{s}'_K)\mat{R}_K^\top\,\bigr].
\label{eq:fura-s}
\end{equation}

Let $\bm{\theta}_t\in\mathbb{R}^d$ collect all block singular values at iteration $t$, initialized at their pretrained values $\bm{\theta}_0$. The population consists of $N/2$ antithetic pairs of perturbed values, $\bm{\theta}_t\pm \mat{D}\mat{z}_j$, with $\mat{z}_j\sim\mathcal{N}(0,\mat{I}_d)$ and $\mat{D}=\operatorname{diag}(\mu_{\mathrm{rel}}|\bm{\theta}_0|)$, so that each singular value moves in proportion to its own size. As in MeZO~\cite{malladi2023mezo}, 
the directions are regenerated from random seeds and never stored. After the $N$ members are scored, their fitness values are standardized as $\widetilde F_j^{\pm}=(F_j^{\pm}-\bar F)/s_F$, where $\bar F$ and $s_F$ are the mean and standard deviation of $\{F_j\}$. The block singular values are updated as
\begin{equation}
\bm{\theta}_{t+1}
=
\bm{\theta}_t+\frac{\alpha}{N}\mat{D}
\sum_{j=1}^{N/2}\bigl(\widetilde F_j^{+}-\widetilde F_j^{-}\bigr)\mat{z}_j,
\label{eq:es-update}
\end{equation}
where $\alpha$ is the step size. Each direction is weighted by the fitness difference of its pair, so the update moves $\bm{\theta}$ toward the perturbations with higher fitness (Figure~\ref{fig:algorithm}b and Algorithm~\ref{alg:furas} in Appendix~\ref{app:setup}). For likelihood fitness, FuRA-S can instead update the block singular values with Adam~\cite{kingma2015adam} on the two-point gradient estimate of MeZO (Appendix~\ref{app:sft}). Since neither update stores activations or gradients, training adds only about 20\,MB to the memory of forward evaluation on Llama-3-8B, even with the Adam moments (Appendix~\ref{app:cost}).

\subsection{Serving specialists on the INT4 generalist}
\label{sec:quant-method}

Phones run models released in quantized form, which store each weight matrix as $\mat{W}_q=(\mat{Q}-\mat{Z})\odot\mat{\Gamma}$ with INT4 codes $\mat{Q}$, zero points $\mat{Z}$, and scales $\mat{\Gamma}$. Storing high-precision block-SVD factors next to these weights would offset the memory saved by quantization, and requantizing $\mat{W}'$ after each iteration would accumulate quantization error. \sys{} therefore rewrites the update of FuRA-S so that the quantized weight itself stays in the computation.

Write the adapted singular values as $\mat{s}'_k=\mat{s}_k\odot(1+\bm{\delta}_k)$. Since $\mat{R}_k^\top \mat{R}_k=\mat{I}$, Eq.~\eqref{eq:block-svd} yields $\mat{L}_k\operatorname{diag}(\mat{s}_k)=\mat{W}_k\mat{R}_k$, and substituting this product into Eq.~\eqref{eq:fura-s} removes $\mat{L}_k$:
\begin{equation}
\mat{W}'=\mat{W}(\mat{I}+\mat{E}),\qquad \mat{E}=\mat{R}\operatorname{diag}(\bm{\delta})\mat{R}^\top,
\label{eq:block-stretch}
\end{equation}
where $\mat{R}=\operatorname{blockdiag}_k(\mat{R}_k)$ and $\bm{\delta}$ concatenates the $\bm{\delta}_k$. This rewrite involves no approximation, because no block singular value is truncated. For the INT4 generalist, $\mat{R}$ comes from the block SVD of its dequantized weights, and each adapted layer computes
\begin{equation}
\mat{y}=\underbrace{\mat{W}_q}_{\text{INT4, frozen}}\Bigl(\mat{x}+\underbrace{\mat{R}}_{\text{frozen}}\operatorname{diag}(\underbrace{\bm{\delta}}_{\text{trainable}})\mat{R}^\top \mat{x}\Bigr).
\label{eq:quant-form}
\end{equation}
In Eq.~\eqref{eq:quant-form}, the large factor $\mat{L}_k$ never appears, and the released $\mat{W}_q$ keeps its integer codes, zero points, and scales, so it runs on the existing INT4 kernels. The frozen $\mat{R}$, which the released checkpoint does not contain, is stored in 4 bits and shared by all specialists, so each specialist stores only its full-precision $\bm{\delta}$ (Appendix~\ref{app:quant-setup}). Since the adaptation only adds $\mat{R}\operatorname{diag}(\bm{\delta})\mat{R}^\top \mat{x}$ to the input, $\bm{\delta}=0$ recovers the quantized model exactly, regardless of the precision of $\mat{R}$.

\subsection{Step-wise fitness for long-horizon tasks}
\label{sec:stepwise}

In long-horizon tasks such as operating a phone app, one wrong action can decide the outcome of an episode, yet the agent receives only a single reward at its end. In MobileWorld~\cite{kong2025mobileworld}, for example, an agent asked to reply to a Mastodon post often searches for the text of the post, which the server does not index, rather than for the account of its author, and then scrolls until it reaches the step limit. When ES scores each member by the rewards of its own episodes, a better action receives credit only if a perturbation replaces the wrong action with it and thereby alters the outcome of the member's few episodes, in which the randomness of all other steps masks its effect.

The step-wise fitness instead scores each action against the whole population. Following GiGPO~\cite{feng2025gigpo}, each step $t$ of an episode with $H$ steps receives the episode reward discounted by $\gamma^{H-1-t}$, with $\gamma\in(0,1)$, as its return. Steps with nearly identical screenshots form an anchor group $g$, in which the credit of step $t$ is the advantage of its action, $\hat Q(g_t,a_t)-\hat V(g_t)$. Here $\hat V(g)$ is the mean return of the steps in $g$, and $\hat Q(g,a)$ is that of the steps taking action $a$. Unlike GiGPO, we pool the groups over all members and sum the credits of each member's steps into its fitness,
\begin{equation}
F_i=\frac{1}{J}\sum_{e=1}^{J}\sum_{t\in e}\bigl[\hat Q(g_t,a_t)-\hat V(g_t)\bigr],
\label{eq:step-fitness}
\end{equation}
where the outer sum runs over the $J$ episodes of member $i$. By the performance-difference lemma~\cite{kakade2002approximately}, $F_i$ approximates the return of member $i$ relative to that of the average policy of the population, whose values $\hat Q$ and $\hat V$ estimate. The step-wise fitness thus ranks the members by return, as the episode rewards do, but assigns the credit to individual actions. This credit is nonzero only on screens where members act differently, since $\hat Q(g,a)=\hat V(g)$ wherever all members take the same action. On these screens, $\hat Q$ pools the episodes of all members, so a member that takes the better action is rewarded even when its own episodes fail for other reasons. The fitness is computed entirely from the screenshots, actions, and rewards that the episodes already record, adding no forward passes or environment calls.

\section{Analyzing FuRA-S through Spectral Inheritance}
\label{sec:method}

\begin{figure*}[t]
\centering
\includegraphics[width=\textwidth]{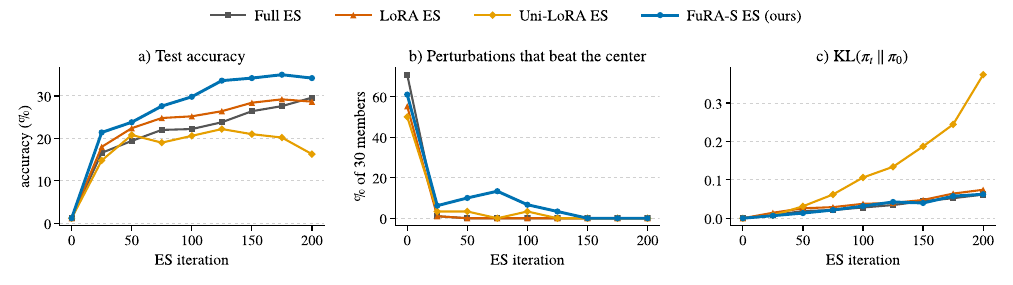}
\caption{ES on Countdown with Qwen2.5-1.5B-Instruct in four search spaces: (a)~accuracy on 500 test problems, (b)~fraction of the 30 perturbations per iteration that improve on the unperturbed model, and (c)~KL divergence from the base policy per response token.}
\label{fig:motivation}
\end{figure*}

This section explains why the search space of FuRA-S suits ES. In existing search spaces, including LoRA adapters and all parameters, ES soon stops finding perturbations that improve the model (Section~\ref{sec:es-stall}). The reason lies in how RLVR changes pretrained weights: the trained weights exhibit spectral inheritance~\cite{zhu2026iso}, and perturbations that respect it improve the model more often (Section~\ref{sec:es-si}). FuRA-S follows this structure with very few parameters, since rescaling the block singular values independently rotates the singular vectors of each weight matrix while barely changing its singular values (Section~\ref{sec:theory}). The weights it trains also exhibit spectral inheritance (Section~\ref{sec:inherit}).

\subsection{ES in existing search spaces}
\label{sec:es-stall}

The search space determines both the size of a specialist and how quickly it improves. LoRA is the natural choice, since it is competitive with full fine-tuning under backpropagation~\cite{schulman2025lora}. Under ES, however, its perturbations soon stop improving the model. We compare four search spaces on the arithmetic reasoning task Countdown with Qwen2.5-1.5B-Instruct, scaling the perturbations of each to the same KL divergence from the unperturbed model. From iteration~50 on, no perturbation of LoRA-ES or full-parameter ES improves on this model (Figure~\ref{fig:motivation}). A smaller search space does not help: Uni-LoRA~\cite{li2025unilora}, which reduces the 4.6M trainable values of LoRA to 508,928, drifts furthest from the base policy and ends with the lowest accuracy. FuRA-S trains as many values but keeps finding improving perturbations through iteration~125 and reaches the highest accuracy. The directions of a search space therefore matter more than its size.

\subsection{ES under spectral inheritance}
\label{sec:es-si}
\label{sec:motivation}

The convergence rate of ES depends on the fraction of perturbations that improve on the unperturbed model. Once this fraction approaches zero, the fitness-weighted update carries little signal and learning slows. An effective search space must therefore keep this fraction positive.

Spectral inheritance, a property of the weight changes that RLVR makes~\cite{zhu2026iso}, points to such a search space. Let $\mat{W}=\mat{U}\mat{\Sigma}\mat{V}^\top$ be the SVD of a pretrained weight matrix $\mat{W}\in\mathbb{R}^{m\times n}$, where $\mat{\Sigma}=\operatorname{diag}(\sigma_1,\dots,\sigma_q)$ holds the spectrum and $q=\min(m,n)$. The left and right singular vectors, the orthonormal columns of $\mat{U}\in\mathbb{R}^{m\times q}$ and $\mat{V}\in\mathbb{R}^{n\times q}$, form the output and input singular frames. The matrices $\mat{U}'\mat{\Sigma}\mat{V}'^\top$ with orthonormal $\mat{U}'$ and $\mat{V}'$ differ from $\mat{W}$ only in their frames and form its \emph{fixed-spectrum family} $\mathcal{F}(\mat{W})$. ISO~\cite{zhu2026iso} finds that RLVR barely moves the spectra of pretrained weights, whereas it rotates both their output and their input frames. Combining the trained frames with the pretrained spectrum retains the gain of RLVR, and optimizing only the frames at a fixed spectrum reaches the end-of-run accuracy of AdamW with 2.7$\times$ fewer training steps on Qwen3-8B-Base. The solutions that RLVR finds therefore lie near the fixed-spectrum family $\mathcal{F}(\mat{W})$.

\begin{figure}[t]
\centering
\includegraphics[width=\columnwidth]{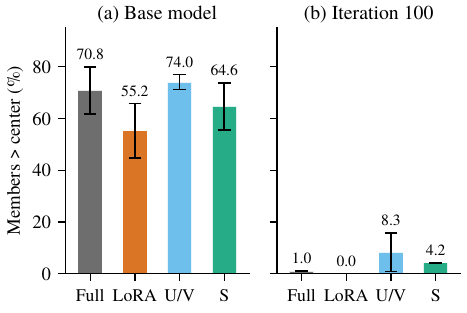}
\caption{Fraction of perturbations that improve on the unperturbed model on Countdown, (a)~around the base model and (b)~around the FuRA-S model at iteration~100.}
\label{fig:beat}
\end{figure}

Perturbations that rotate the frames at a fixed spectrum should improve on the unperturbed model more often, even late in training. We test this on Countdown with Qwen2.5-1.5B-Instruct over three seeds, comparing four kinds of perturbations matched in KL divergence, which change all weights (Full), rank-4 LoRA adapters (LoRA), only the frames at a fixed $\mat{\Sigma}$ (U/V), or only the global singular values in $\mat{\Sigma}$ (S). Around the base model, frame-only perturbations improve on it slightly more often than those of all weights, 74.0\% compared with 70.8\% (Figure~\ref{fig:beat}a). Around the model of FuRA-S at iteration~100, 8.3\% of them still do, compared with 1.0\% for Full and none for LoRA (Figure~\ref{fig:beat}b).

\observation{ES finds improving perturbations more often by rotating the frames of each weight matrix at a fixed spectrum, where the solutions of RLVR lie.}

Rotating the frames through the SVD of each full weight matrix, however, is impractical on a phone. A specialist that changes both $\mat{U}$ and $\mat{V}$ would be larger than the model itself, and for INT4 weights these dense factors would also need high-precision storage. Smaller alternatives fall short as well: low-rank adapters initialized from a truncated SVD suffer from spectral collapse in RLVR~\cite{yin2025perlvr}, and perturbing only the global singular values, as in SVF~\cite{sun2022svf}, cannot rotate the frames. The block singular values that FuRA-S trains avoid these problems, and Section~\ref{sec:theory} shows that perturbing them still rotates the frames.

\subsection{How block rescaling follows spectral inheritance}
\label{sec:theory}
\label{sec:analysis}

We now show that FuRA-S follows spectral inheritance: its perturbations rotate the frames of $\mat{W}$ while keeping $\mat{W}'$ near $\mathcal{F}(\mat{W})$. As in ISO~\cite{zhu2026iso}, we quantify how much an update $\Delta \mat{W}=\mat{W}'-\mat{W}$ changes the spectrum with two metrics (Appendix~\ref{app:linearization}). The \emph{spectral residual} $\rho_\Sigma$ is the distance from $\mat{W}'$ to $\mathcal{F}(\mat{W})$ relative to the update size, ranging from 0 for a pure rotation to 1 for a pure rescaling of singular values. The \emph{spectrum-changing energy} $\kappa$ is the fraction of the update energy that changes the singular values to first order, normalized to one for a random dense update. We measure frame rotation by the principal angles between the leading singular subspaces of $\mat{W}$ and $\mat{W}'$.

\paragraph{Frame rotation from block rescaling.}
By Eq.~\eqref{eq:block-stretch}, rescaling each $\mat{s}_k$ by $1+\bm{\delta}_k$ gives $\mat{W}'=\mat{W}(\mat{I}+\mat{E})$ with $\mat{E}=\mat{R}\operatorname{diag}(\bm{\delta})\mat{R}^\top$. In the singular basis of $\mat{W}$, the update decomposes as $\mat{U}^\top\Delta \mat{W}\mat{V}=\mat{\Sigma} \mat{G}$ with $\mat{G}=\mat{V}^\top \mat{E}\mat{V}$: the diagonal of $\mat{G}$ changes the spectrum, and its off-diagonal entries rotate both frames (Proposition~\ref{prop:block}, Appendix~\ref{app:linearization}). Rescaling the singular values of each block \emph{independently} shifts the relative contributions of the blocks to the assembled matrix, producing off-diagonal entries in $\mat{G}$ and thereby rotating $\mat{U}$ and $\mat{V}$. With a single block, the block SVD is the SVD of $\mat{W}$, so $\mat{R}=\mat{V}$, $\mat{G}$ is diagonal, and the same rescaling changes only the spectrum. The rotation therefore arises because each global singular vector spans several blocks, whose singular values are rescaled differently.

Training only the block singular vectors, by contrast, is less effective at rotating the frames. Rotating $\mat{R}_k$ leaves $\mat{U}$ and $\mat{\Sigma}$ unchanged and rotates only $\mat{V}$, whereas RLVR rotates both frames~\cite{zhu2026iso}. Rotating $\mat{L}_k$ within its column span limits how far $\mat{U}$ can move. Dropping the orthogonality of $\mat{R}_k$ rotates $\mat{U}$ as well but turns each $\mat{R}_k$ into a dense factor with $b$ times as many values as $\mat{s}_k$, losing the parameter efficiency that on-device training requires.

\paragraph{Spectrum preservation.}
To first order, the relative change of each singular value is a weighted average of the relative changes of the block singular values (Proposition~\ref{prop:block}):
\begin{equation}
\frac{\Delta\sigma_i}{\sigma_i}=\sum_{j=1}^{n}w_{ij}\,\delta_j,\qquad w_{ij}=(\mat{r}_j^\top \mat{v}_i)^2,
\label{eq:weighted-avg}
\end{equation}
where $\mat{r}_j$ are the columns of $\mat{R}$ and the weights $w_{ij}$ are nonnegative and sum to one over $j$. Since ES perturbs the block singular values independently with zero mean, the positive and negative $\delta_j$ from different blocks largely cancel in this average, and the spectrum barely moves.

\paragraph{Measured rotation.}
We verify both properties by applying random updates at the scale of one ES perturbation, $\|\Delta \mat{W}\|_F=0.03\|\mat{W}\|_F$, to each block-SVD factor of 12 Qwen2.5-1.5B-Instruct matrices (Table~\ref{tab:core-choice} in Appendix~\ref{app:diagnostics}). All updates preserve the spectrum ($\rho_\Sigma\le 5.3\%$, $\kappa\le 7.5$), far below the values ISO reports for supervised fine-tuning, but differ sharply in frame rotation. Rescaling $\mat{s}_k$ rotates $\mat{U}$ and $\mat{V}$ by about $9^\circ$, four times as far as a dense update of $\mat{W}$, whereas every block singular-vector update rotates them less despite $(b{-}1)/2$ to $b$ times as many values. Rescaling the global singular values rotates neither frame (Figure~\ref{fig:diagnostics}a,b in Appendix~\ref{app:diagnostics}).

\takeaway{Perturbing only the block singular values preserves the spectrum of each weight matrix but rotates both of its frames four times as far as a dense update.}

\subsection{Spectral inheritance of the trained weights}
\label{sec:inherit}

Individual perturbations follow spectral inheritance (Section~\ref{sec:theory}), but the accumulated update, a fitness-weighted sum of many such perturbations, need not preserve this structure. We verify that it does by decomposing the weights that FuRA-S trains on MATH (Qwen2.5-Math-7B) and Countdown (Qwen2.5-1.5B-Instruct) into rotated frames and a changed spectrum. The accuracy gain comes almost entirely from the rotated frames: combined with the pretrained spectrum, they retain 95.1\% of it on MATH and 95.8\% on Countdown (Figure~\ref{fig:algorithm}c). Conversely, the trained spectrum with the pretrained frames retains only 15\% of the gain on MATH and none on Countdown. The trained weights thus exhibit spectral inheritance, as RLVR-trained models do~\cite{zhu2026iso}.

\takeaway{Accumulated over training, the updates of FuRA-S still follow spectral inheritance: the rotated frames alone retain over 95\% of the accuracy gain.}

\begin{table*}[t]
\caption{Countdown accuracy (\%) on 2,000 test problems. The BP-based policy-gradient baselines are from ES at Scale, Full ES is our reproduction of ES at Scale (Appendix~\ref{app:countdown-setup}), and LoRA-ES uses rank-4 adapters, and VeRA-ES uses rank-1024 VeRA adapters. Best per row in bold, and parentheses give the difference from Full ES.}
\vspace{5pt}
\label{tab:countdown}
\centering
\begin{tabular*}{\textwidth}{@{\extracolsep{\fill}}lrrrrrrrr}
\toprule
 & & \multicolumn{3}{c}{BP-based} & \multicolumn{4}{c}{BP-free} \\
\cmidrule(lr){3-5}\cmidrule(lr){6-9}
Base model & Original & PPO & GRPO & Dr.~GRPO & Full ES & LoRA-ES & VeRA-ES & FuRA-S (ours) \\
\midrule
Qwen2.5-0.5B-Instruct & 0.1 & 0.3 & 13.0 & 13.5 & \textbf{14.4} & 12.6 & 13.5 & \textbf{14.4} (0.0) \\
Llama-3.2-1B-Instruct & 0.4 & 11.2 & 14.9 & 13.9 & 15.6 & 11.7 & 13.7 & \textbf{15.8} \posd{+0.2} \\
Qwen2.5-1.5B-Instruct & 0.7 & 14.2 & 27.8 & 31.0 & 36.7 & 34.7 & 28.1 & \textbf{37.0} \posd{+0.3} \\
Qwen2.5-3B-Instruct & 10.0 & 20.1 & 37.8 & 43.8 & \textbf{58.2} & 54.3 & 55.2 & 55.8 \negd{$-$2.4} \\
Qwen2.5-7B-Instruct & 31.2 & 55.1 & 57.0 & 57.5 & \textbf{58.9} & 55.5 & 54.2 & 56.2 \negd{$-$2.7} \\
\bottomrule
\end{tabular*}
\end{table*}

\begin{table*}[t]
\caption{Greedy pass@1 (\%) of Qwen2.5-Math-7B at the final checkpoint, with FuRA-S minus Full ES in parentheses.}
\vspace{5pt}
\label{tab:math}
\centering
\setlength{\tabcolsep}{9pt}
\begin{tabularx}{\textwidth}{>{\hspace*{-\tabcolsep}}Xrrrrrrr}
\toprule
Method & AIME24 & AMC & MATH500 & Minerva & Olympiad & Avg-5 & Avg-3 \\
\midrule
Full ES & 26.7 & 57.8 & 77.6 & 26.8 & 37.0 & \textbf{45.2} & 47.2 \\
LoRA-ES & 16.7 & 53.0 & 74.0 & 22.8 & 35.4 & 40.4 & 44.1 \\
\rowcolor{oursbg} FuRA-S (ours) & 23.3 & 54.2 & 76.8 & 26.5 & 40.0 & 44.2 \negd{$-$1.0} & \textbf{47.8} \posd{+0.6} \\
\bottomrule
\end{tabularx}
\end{table*}

\section{Algorithm Evaluation}
\label{sec:experiments}

We first evaluate FuRA-S on standard benchmarks, testing whether it matches full-parameter ES with a tiny fraction of the parameters (Section~\ref{sec:rlvr}), outperforms ES over LoRA adapters and quantized weights (Sections~\ref{sec:rlvr} and~\ref{sec:quant}), and remains effective beyond single-turn RLVR (Section~\ref{sec:beyond}).

\subsection{RLVR on Countdown and MATH}
\label{sec:rlvr}

Countdown and MATH are scored by automatic checks of the final answer, the first kind of fitness of Section~\ref{sec:workflow}. On both tasks, we compare FuRA-S with full-parameter ES and with LoRA-ES on rank-4 adapters, which uses the same number of members and per-matrix perturbation size as FuRA-S, so that only the search space differs. On Countdown, we also compare with the policy-gradient methods PPO~\cite{schulman2017ppo}, GRPO~\cite{shao2024deepseekmath}, and Dr.~GRPO~\cite{liu2025drgrpo}, which train all parameters by backpropagation (BP).

\paragraph{Countdown.}
\label{sec:countdown}
Following ES at Scale~\cite{qiu2025es}, we train five models from 0.5B to 7B parameters on 200 Countdown problems (Appendix~\ref{app:countdown-setup}). FuRA-S is on par with full-parameter ES, matching or exceeding it on the three smaller models and staying within three points on the two larger ones (Table~\ref{tab:countdown}), although it trains at most 0.05\% of the parameters that full-parameter ES perturbs. It also outperforms LoRA-ES on every model while training about nine times fewer parameters, which reflects the advantage of a search space that follows spectral inheritance (Section~\ref{sec:theory}). VeRA-ES, which perturbs VeRA adapters~\cite{kopiczko2023vera} with only 1.4 to 1.9 times as many trainable parameters as FuRA-S, falls below FuRA-S on every model as well, by up to 8.9 points on Qwen2.5-1.5B, so this advantage does not come from the smaller search space alone. Finally, without computing any gradient, FuRA-S surpasses all BP-based baselines on four of the five models, with the largest margin on Qwen2.5-3B.

\paragraph{MATH.}
\label{sec:math}
We further train Qwen2.5-Math-7B under the protocol of ES at Scale and evaluate it on five mathematics benchmarks (Appendix~\ref{app:math-setup}). The comparisons with ES carry over to this harder task (Table~\ref{tab:math}). FuRA-S slightly exceeds full-parameter ES on the average over the three larger benchmarks (Avg-3) and stays within one point on the average over all five (Avg-5), which also includes the small AIME24 and AMC sets. On both averages, it again outperforms LoRA-ES at the final checkpoint and at the best checkpoints by Avg-5 and by Avg-3 (Table~\ref{tab:math-full}).

\subsection{RLVR on quantized models}
\label{sec:quant}

Since phones run quantized models, we adapt the released GPTQ-INT4~\cite{frantar2023gptq} checkpoints of Qwen2.5-1.5B-Instruct and Qwen2.5-3B-Instruct on Countdown while keeping their 4-bit weights unchanged. We compare FuRA-S with QES~\cite{xu2026qes}, which applies full-parameter ES to quantized weights, under its original protocol. Searching the block singular values instead of the quantized weights is more effective on both models. \mbox{FuRA-S} outperforms our reproduction of QES by 5.85 and 16.25 points after 300 iterations and exceeds the best accuracy of the entire QES run within the first 50 iterations (Figure~\ref{fig:quant} in Appendix~\ref{app:quant-setup}). Both ES methods also far outperform QuZO, the zeroth-order baseline of \citet{xu2026qes}. Because a specialist leaves the INT4 weights intact, serving it adds only about 9\% memory and 11--15\% decoding time per token on the existing INT4 kernels (Appendix~\ref{app:cost}).

\subsection{Supervised and long-horizon objectives}
\label{sec:beyond}

We also examine whether the search space supports supervised fine-tuning and RLVR over many turns. Supervised fine-tuning maximizes the likelihood of the data, as \sys{} does with the user's records, but it changes the spectrum much more than RLVR does (Figure~\ref{fig:diagnostics}c,d in Appendix~\ref{app:diagnostics}). On Commonsense-170K~\cite{hu2023llmadapters} with Llama-3-8B, FuRA-S outperforms all other parameter-efficient forward-only methods under the same budget of forward passes and exceeds the strongest of them, VeRA, by about three points (Appendix~\ref{app:sft}). The search space thus remains effective for an objective whose weight changes do not follow spectral inheritance. The agentic Sudoku benchmark of Agentic ESOpt~\cite{zheng2026agenticesopt} tests RLVR over many turns, since the agent acts for up to 45 turns and receives a single reward at the end of the episode, as in phone-use tasks. On this benchmark, FuRA-S matches the held-out success that Agentic ESOpt reports for full-parameter ES with a constant perturbation scale (Appendix~\ref{app:sudoku}).

\subsection{Ablations}
\label{sec:ablation}

We ablate the block partition and the perturbation distribution on Countdown with Qwen2.5-3B-Instruct, where FuRA-S reaches 49.25\% accuracy after 200 iterations. Replacing the block SVD with the SVD of the whole matrix reduces FuRA-S to ES over the global singular values, the search space of SVF. Its perturbations change only the spectrum and rotate neither frame (Section~\ref{sec:theory}), and accuracy drops by about four points to 45.40\%. Frame rotation, which only the block structure provides, therefore contributes to the gain of FuRA-S. Replacing the Gaussian perturbations with Rademacher ones, which change every block singular value by exactly $\pm\mu_{\mathrm{rel}}$ times its pretrained value, reduces accuracy by almost ten points to 39.60\%. Since Rademacher perturbations have the zero mean and equal variance that the first-order analysis assumes (Corollary~\ref{cor:block} in Appendix~\ref{app:linearization}), the benefit of Gaussian perturbations comes from properties beyond these two moments.

\takeaway{FuRA-S matches full-parameter ES with at most 0.05\% of the parameters, outperforms ES over LoRA adapters or quantized weights, and surpasses BP-based policy-gradient methods on most models.}

\section{On-Device Training Evaluation}
\label{sec:device}

We next evaluate \sys{} in its target settings, testing whether the step-wise fitness makes long-horizon phone-use tasks trainable (Section~\ref{sec:gui}), whether specialists trained on the likelihood of the user's records compete with per-user LoRA trained by backpropagation (Section~\ref{sec:lamp}), and whether training fits in the memory of a phone (Section~\ref{sec:device-iphone}).

\subsection{Phone-use agents}
\label{sec:gui}

\begin{table}[t]
\caption{Success rate (\%) of MAI-UI-2B on MobileWorld tasks, for the generalist and its specialists after up to 15 ES iterations.}
\vspace{5pt}
\label{tab:gui}
\centering
\small
\setlength{\tabcolsep}{4pt}
\resizebox{\columnwidth}{!}{%
\begin{tabular}{lrrr}
\toprule
Task & Base & Episode reward & Step-wise (ours) \\
\midrule
MastodonReply & 23.4 & 79.7 & \textbf{89.1} \\
CheckEventTime & 9.4 & 92.2 & \textbf{100.0} \\
RequestCarpooling & 6.2 & 14.1 & \textbf{98.4} \\
MastodonFollow & 23.4 & \textbf{64.1} & \textbf{64.1} \\
MastodonPostPoll & 9.4 & 20.3 & \textbf{68.8} \\
\bottomrule
\end{tabular}}
\end{table}

\begin{figure*}[t]
\centering
\includegraphics[width=\textwidth]{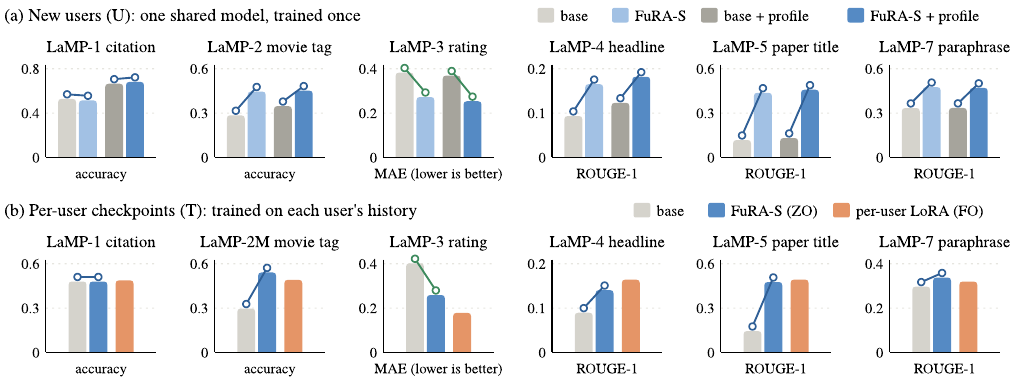}
\caption{LaMP with Llama-3.1-8B-Instruct. (a)~User-based setting, without and with the retrieved profile. (b)~Time-based setting, with per-user specialists trained by ES (ZO in the legend) and per-user LoRA trained by backpropagation (FO in the legend).}
\label{fig:lamp}
\end{figure*}

Training a phone-use agent for its user's tasks is the main use case of \sys{}. We evaluate this use case on the GUI-only tasks of the MobileWorld benchmark~\cite{kong2025mobileworld}, which involve real Android apps such as Mastodon, an email client, and a calendar. In each task, the agent has up to 30 steps to accomplish a natural-language goal on a freshly reset phone, after which a rule-based checker inspects the device and returns the reward (Appendix~\ref{app:gui}).

MAI-UI-2B~\cite{zhou2025maiui} succeeds on only 11.1\% of these tasks, and online reinforcement learning raises the success rate to 17.1\%, as reported by \citet{tang2026clawgui}. A user, however, needs the agent to succeed on the tasks that they perform, so \sys{} trains one specialist for each of them. We train such specialists on five tasks on which MAI-UI-2B succeeds in fewer than a quarter of the episodes. The tasks are to reply to a Mastodon post (MastodonReply), set an alarm from the time of an event in an email (CheckEventTime), text a contact for a ride (RequestCarpooling), follow a given Mastodon account (MastodonFollow), and post a multiple-choice poll with given hashtags (MastodonPostPoll). Each specialist changes only the 516,096 block singular values of the model, 0.024\% of its parameters, and trains for 15 iterations, or 960 episodes. We compare the step-wise fitness with episode rewards (Section~\ref{sec:stepwise}).

The step-wise fitness turns tasks that the generalist rarely solves into tasks that its specialists solve reliably (Table~\ref{tab:gui}). The specialists succeed in nearly every episode on three tasks and improve substantially on the other two. Episode rewards also improve on the generalist but trail the step-wise fitness on four of the five tasks and fail almost entirely on RequestCarpooling, since an episode reward cannot tell which action moved the member toward the goal.

\subsection{Personalization}
\label{sec:lamp}

\begin{figure}[t]
\centering
\includegraphics[width=\columnwidth]{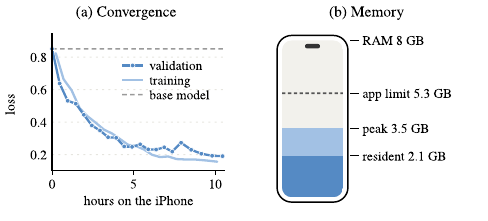}
\caption{Training a specialist for a LaMP-2M user on an iPhone 16 Pro: (a)~training and validation loss, (b)~memory use.}
\label{fig:device}
\end{figure}

Personalization tests the second kind of fitness of Section~\ref{sec:workflow}, the likelihood of the user's own records. We evaluate \sys{} on six LaMP tasks~\cite{salemi2024lamp} with Llama-3.1-8B-Instruct in the user-based setting of LaMP and the time-based setting of OPPU~\cite{tan2024oppu}. In the user-based setting (U), one specialist per task is trained on some users and tested on unseen users, whose prompts either omit or include one record retrieved from their profile of past records. In the time-based setting (T), a separate specialist is trained for each user on their earlier records and tested on later queries, as \sys{} does on the phone. We compare these specialists with per-user LoRA trained by backpropagation as in OPPU (Appendix~\ref{app:lamp-setup}).

The likelihood of the user's records proves to be an effective fitness function. In the user-based setting, the specialists improve over the generalist on all six tasks when both use the retrieved profile, with the largest gain on the title-generation task LaMP-5 (Figure~\ref{fig:lamp}a and Table~\ref{tab:lamp-u-llama}). In the time-based setting, the per-user specialists nearly double the accuracy of the generalist on the movie-tagging task LaMP-2M and more than triple its ROUGE-1 on LaMP-5. With forward passes alone, they outperform per-user LoRA trained by backpropagation on LaMP-2M and the tweet-paraphrasing task LaMP-7 (Figure~\ref{fig:lamp}b and Table~\ref{tab:lamp-t-llama}), while each specialist stores only its 1.25M relative changes $\bm{\delta}$, or 4.98\,MB. The gains in the user-based setting hold for Qwen3.5-4B on all six tasks, and its per-user specialists also outperform per-user LoRA on LaMP-1 and LaMP-2M (Appendix~\ref{app:lamp-qwen}).

\subsection{Training on an iPhone}
\label{sec:device-iphone}

We run \sys{} on an iPhone 16 Pro with 8\,GB of memory and train a specialist for one LaMP-2M user on top of the released GPTQ-INT4~\cite{frantar2023gptq} Llama-3.1-8B-Instruct, following the time-based protocol of Section~\ref{sec:lamp}. The phone reproduces the forward computation of a GPU server, with a validation loss of 0.8505 for the generalist on the phone and 0.8524 on the server. After 20 iterations over 10.4 hours, the validation loss drops from 0.85 to 0.19 (Figure~\ref{fig:device}a), and the specialist answers 8 of 15 held-out queries correctly, compared with 5 for the generalist. The INT4 model occupies 5.5\,GB, more than the 5.3\,GB available to an app, but since \sys{} streams the decoder layers from flash, training uses only 2.1\,GB of resident memory and peaks at 3.5\,GB (Figure~\ref{fig:device}b). Forward computation accounts for 97\% of the training time, and streaming takes only 14 seconds per iteration. Training a specialist for an 8B model thus finishes overnight within the memory available to an app.

\takeaway{\sys{} trains specialists that reliably solve phone-use tasks on which the generalist rarely succeeds, and it trains per-user specialists of a few megabytes within the memory of a phone.}

\section{Conclusion}
\label{sec:conclusion}

We have presented \sys{}, an on-device system that trains specialist LLMs with forward passes over the INT4 model that the phone already runs. At its core, FuRA-S trains only the block-spectral coordinates of pretrained weights, whose independent rescaling rotates the singular vectors of each weight matrix while largely preserving its singular values, as RLVR does. With specialists of 1--5\,MB, \sys{} raises the success rate of a phone-use agent from 6--23\% up to 89--100\% and trains a personalized 8B model on an iPhone overnight. Training with forward passes only remains slow, since each iteration evaluates a population of perturbed models and a multi-step task multiplies this cost by the episodes of every member and the steps of every episode. Each task also needs a fitness that the device can compute without labels, specialists do not yet share what they learn, and our evaluation covers five MobileWorld tasks and six LaMP tasks. More broadly, training specialist LLMs on personal devices becomes practical once the search space is small enough to store for every task and aligned with how RLVR changes pretrained weights, so that a phone can adapt a model to its user while the user's data never leaves the device.

\clearpage
\bibliography{example_paper,agent_extra}
\bibliographystyle{mlsys2025}

\clearpage
\onecolumn
\raggedbottom
\setlength{\intextsep}{4pt}
\appendix
\label{appendix-start}
\setcounter{table}{0}
\renewcommand{\thetable}{A\arabic{table}}

\section{First-Order Effect of Block Rescaling}
\label{app:linearization}

Section~\ref{sec:theory} uses the following two results.

\begin{proposition}[Block rescaling]
\label{prop:block}
Suppose $\mat{W}$ has distinct positive singular values. To first order in $\bm{\delta}$, the relative change of each singular value of $\mat{W}$ is the weighted average of the relative changes of the block singular values in Eq.~\eqref{eq:weighted-avg}, with nonnegative weights $w_{ij}$ that sum to one over $j$. Each off-diagonal entry $G_{ik}$ rotates the singular vectors $\mat{u}_k$ and $\mat{v}_k$ toward $\mat{u}_i$ and $\mat{v}_i$, with a magnitude that grows as $\sigma_i$ approaches $\sigma_k$.
\end{proposition}

\begin{corollary}[Source of the rotation]
\label{cor:block}
(i)~If all $\delta_j$ equal $c$, then $\mat{W}'=(1+c)\mat{W}$, which scales the spectrum and leaves the singular vectors unchanged. (ii)~If $\mat{W}$ forms a single block ($K=1$), then $\mat{R}=\mat{V}$ and $\mat{G}=\operatorname{diag}(\bm{\delta})$, so the update changes only the spectrum. (iii)~If $m\ge n$ and the entries of $\bm{\delta}$ are independent with zero mean and equal variance, the diagonal of $\mat{G}$ accounts for an expected fraction $\frac{1}{n}\sum_{i,j}w_{ij}^2$ of $\mathbb{E}\|\mat{G}\|_F^2$, and the off-diagonal entries account for the rest.
\end{corollary}

\noindent We prove both below. Let $\mat{W}=\mat{U}\mat{\Sigma} \mat{V}^\top$ be the reduced SVD of $\mat{W}\in\mathbb{R}^{m\times n}$ with distinct positive singular values, and let $\mat{W}'=\mat{W}(\mat{I}+\mat{E})$ with $\mat{E}=\mat{R}\operatorname{diag}(\bm{\delta})\mat{R}^\top$ as in Eq.~\eqref{eq:block-stretch}. The update is $\Delta \mat{W}=\mat{W}'-\mat{W}=\mat{U}\mat{\Sigma} \mat{V}^\top \mat{E}$, so that $\mat{U}^\top\Delta \mat{W} \mat{V}=\mat{\Sigma} \mat{G}$ with $\mat{G}=\mat{V}^\top \mat{E}\mat{V}$. We write $\mat{E}=\sum_{j}\delta_j \mat{r}_j\mat{r}_j^\top$, where $\mat{r}_j$ is the $j$-th column of the orthogonal matrix $\mat{R}$.

\paragraph{Block identity.}
Since $\mat{R}_k$ is orthogonal, $\mat{L}_k\operatorname{diag}(\mat{s}_k\odot(1+\bm{\delta}_k))\mat{R}_k^\top=\mat{L}_k\operatorname{diag}(\mat{s}_k)\mat{R}_k^\top \mat{R}_k\operatorname{diag}(1+\bm{\delta}_k)\mat{R}_k^\top=\mat{W}_k\bigl(\mat{I}+\mat{R}_k\operatorname{diag}(\bm{\delta}_k)\mat{R}_k^\top\bigr)$. Stacking the blocks gives Eq.~\eqref{eq:block-stretch}.

\paragraph{Singular values.}
For a small update, first-order perturbation theory gives $\Delta\sigma_i\approx \mat{u}_i^\top\Delta \mat{W} \mat{v}_i$ for each simple singular value. Since $\mat{u}_i^\top \mat{U}\mat{\Sigma} \mat{V}^\top=\sigma_i \mat{v}_i^\top$, this equals $\sigma_i \mat{v}_i^\top \mat{E} \mat{v}_i=\sigma_i G_{ii}$, and
\begin{equation}
G_{ii}=\sum_{j=1}^{n}\delta_j\,(\mat{r}_j^\top \mat{v}_i)^2=\sum_{j=1}^{n}w_{ij}\,\delta_j .
\end{equation}
The weights $w_{ij}$ are nonnegative and sum to $\|\mat{R}^\top \mat{v}_i\|_2^2=\|\mat{v}_i\|_2^2=1$, since $\mat{R}$ is orthogonal. This part holds for any shape of $\mat{W}$.

\paragraph{Singular vectors.}
The same theory gives the first-order changes of the singular vectors:
\begin{align}
\mat{v}'_k-\mat{v}_k &\approx \sum_{i\ne k}\frac{\sigma_i^2+\sigma_k^2}{\sigma_k^2-\sigma_i^2}\,G_{ik}\,\mat{v}_i, \nonumber\\
\mat{u}'_k-\mat{u}_k &\approx \sum_{i\ne k}\frac{2\sigma_i\sigma_k}{\sigma_k^2-\sigma_i^2}\,G_{ik}\,\mat{u}_i.
\label{eq:rotation}
\end{align}
The off-diagonal entries of $\mat{G}$ therefore rotate the singular vectors, with coefficients that grow as $\sigma_i$ approaches $\sigma_k$. When $\mat{W}$ has more columns than rows and full row rank, the right singular vectors additionally move outside the original row space, by
\begin{equation}
(\mat{I}-\mat{V}\mat{V}^\top)\mat{E}\mat{v}_k
\label{eq:rotation-null}
\end{equation}
to first order. These expressions assume distinct singular values. For repeated singular values, the perturbation must instead be characterized at the level of singular subspaces.

\paragraph{Corollary~\ref{cor:block}.}
(i)~If $\delta_j=c$ for all $j$, then $\mat{E}=c\mat{R}\mat{R}^\top=c\mat{I}$, and $\mat{W}'=(1+c)\mat{W}$ has the singular vectors of $\mat{W}$ and the spectrum $(1+c)\mat{\Sigma}$. (ii)~If $K=1$, the block SVD is the SVD of $\mat{W}$, so $\mat{R}=\mat{V}$, with the reduced factors when $m<n$, and $\mat{G}=\operatorname{diag}(\bm{\delta})$. All off-diagonal entries of $\mat{G}$ then vanish, and so does the term in Eq.~\eqref{eq:rotation-null}, since $\mat{E}\mat{v}_k=\delta_k\mat{v}_k$ lies in the row space. (iii)~Let $m\ge n$, so that $\mat{V}$ is square and orthogonal, and let the entries of $\bm{\delta}$ be independent with zero mean and variance $\nu^2$. Then $\mathbb{E}\|\mat{G}\|_F^2=\mathbb{E}\|\mat{E}\|_F^2=\mathbb{E}\|\bm{\delta}\|_2^2=n\nu^2$, while
\begin{equation}
\mathbb{E}\,G_{ii}^2=\sum_{j,l}w_{ij}w_{il}\,\mathbb{E}[\delta_j\delta_l]=\nu^2\sum_{j}w_{ij}^2 .
\end{equation}
The diagonal therefore holds the expected fraction $\frac{1}{n}\sum_{i,j}w_{ij}^2$ of $\mathbb{E}\|\mat{G}\|_F^2$. Since the weights of each $\mat{v}_i$ sum to one, $\sum_j w_{ij}^2$ lies between $1/n$, when $\mat{v}_i$ spreads evenly over all columns of $\mat{R}$, and 1, when $\mat{v}_i$ is a column of $\mat{R}$.

\paragraph{Spectral metrics.}
Section~\ref{sec:theory} and Appendix~\ref{app:diagnostics} use the metrics of ISO~\cite{zhu2026iso},
\begin{equation}
\begin{aligned}
\rho_\Sigma&=\frac{\|\sigma(\mat{W}')-\sigma(\mat{W})\|_2}{\|\Delta \mat{W}\|_F},\\
\kappa&=\frac{mn}{q}\,\frac{\|\operatorname{diag}(\mat{U}^\top\Delta \mat{W} \mat{V})\|_2^2}{\|\Delta \mat{W}\|_F^2},
\end{aligned}
\label{eq:spectral-metrics}
\end{equation}
where $\sigma(\cdot)$ denotes the sorted singular values. Since $\mat{W}\in\mathcal{F}(\mat{W})$, the distance from $\mat{W}'$ to $\mathcal{F}(\mat{W})$ is at most $\|\Delta \mat{W}\|_F$, so $0\le\rho_\Sigma\le1$. The matrices $\mat{u}_i\mat{v}_i^\top$ form an orthonormal basis of the $q$-dimensional subspace of first-order spectrum-changing directions, and an isotropic random update of fixed Frobenius norm places an expected fraction $q/(mn)$ of its squared norm in this subspace, so $\kappa$ has expectation one. Since $mn/q=\max(m,n)$, $\kappa$ equals $\max(m,n)$ times the share of $\|\Delta \mat{W}\|_F^2$ on the diagonal of $\mat{U}^\top\Delta \mat{W} \mat{V}$.

\section{Measured Rotation and Spectral Metrics}
\label{app:diagnostics}

\begin{figure}[H]
\centering
\includegraphics[width=\textwidth,trim=0 0 0 158.3bp,clip]{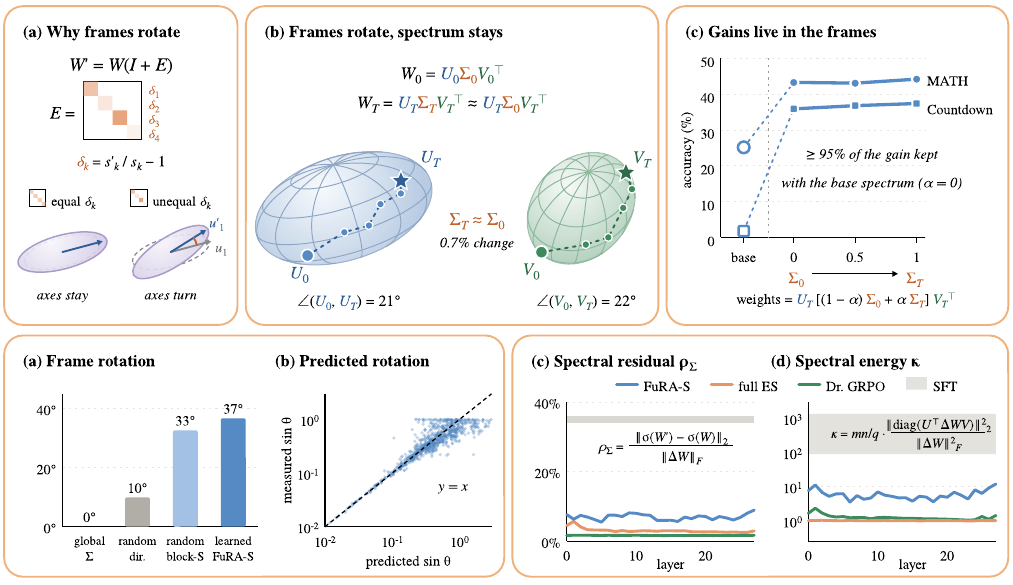}
\caption{(a)~Rotation of the leading singular vectors of eight Qwen2.5-1.5B-Instruct matrices under four updates with the norm of the update learned on Countdown, and (b)~its first-order prediction at small sizes. (c,\,d)~Per-layer $\rho_\Sigma$ and $\kappa$ of Qwen2.5-Math-7B after MATH training, with the values reported by ISO for supervised fine-tuning in gray.}
\label{fig:diagnostics}
\end{figure}

Table~\ref{tab:core-choice} lists the random updates of the block-SVD factors that Section~\ref{sec:theory} compares. Figure~\ref{fig:diagnostics}(a) measures the rotation of the leading singular vectors of eight Qwen2.5-1.5B-Instruct matrices under four updates with the norm of the update learned on Countdown, and panel (b) compares it with its first-order prediction at small update sizes. At the size of the update learned on Countdown, a random block rescaling rotates the leading singular vectors by $33^\circ$ and the learned one by $37^\circ$, compared with $10^\circ$ for a dense update. Panels (c) and (d) give the per-layer $\rho_\Sigma$ and $\kappa$ of Qwen2.5-Math-7B after training on MATH, together with the values that ISO~\cite{zhu2026iso} reports for supervised fine-tuning.

\begin{table}[H]
\caption{Random updates of different block-SVD factors (medians). $\angle \mat{U}$, $\angle \mat{V}$: largest principal angles between the leading 16-dimensional singular subspaces of $\mat{W}$ and $\mat{W}'$.}
\label{tab:core-choice}
\centering
\small
\setlength{\tabcolsep}{3.5pt}
\begin{tabular}{lrrrrr}
\toprule
Updated factor & Values & $\rho_\Sigma$ & $\kappa$ & $\angle \mat{U}$ & $\angle \mat{V}$\\
\midrule
\rowcolor{oursbg} Block $\mat{s}_k$ (FuRA-S) & 1.5k & 5.3\% & 7.5 & 8.6$^\circ$ & 8.8$^\circ$\\
$\mat{R}_k$, additive & 135k & 3.0\% & 1.2 & 6.1$^\circ$ & 6.4$^\circ$\\
$\mat{R}_k$, rotation & 67k & 0.0\% & 1.2 & 0.0$^\circ$ & 1.9$^\circ$\\
$\mat{L}_k$, rotation in span & 67k & 0.6\% & 1.4 & 1.8$^\circ$ & 2.4$^\circ$\\
Dense $\mat{W}$ & 8M & 2.3\% & 1.0 & 2.1$^\circ$ & 2.1$^\circ$\\
\bottomrule
\end{tabular}
\end{table}

\section{Experimental Details}
\label{app:setup}

Table~\ref{tab:hparams} summarizes the hyperparameters of the experiments in Sections~\ref{sec:experiments} and~\ref{sec:device}, including the supervised and agentic experiments of Appendices~\ref{app:sft} and~\ref{app:sudoku}. Unless otherwise specified, FuRA-S adapts all linear projections in the decoder layers using the block partition of Section~\ref{sec:coords}, in which $n=Kb$ and $K\le b$ are the factor pair of $n$ closest to $\sqrt{n}$, as in FuRA, and attention and gated-DeltaNet output projections are split into blocks by head. For $m<b$, the reduced SVD of a block contains $m$ singular values. The block SVD of Qwen2.5-1.5B takes about 32 seconds on one A100 GPU. Perturbations are scaled by the pretrained singular values, and reward-based ES standardizes fitness across each population without momentum.

Algorithm~\ref{alg:furas} summarizes the ES loop of FuRA-S (Section~\ref{sec:optim}).

\begin{algorithm}[H]
\caption{FuRA-S: forward-only block-spectral ES}
\label{alg:furas}
\centering
\begin{minipage}{0.6\textwidth}
\small
\begin{algorithmic}[1]
\REQUIRE Pretrained weights $\{\mat{W}\}$, perturbation scale $\mu_{\mathrm{rel}}$, step size $\alpha$, population $N$, iterations $T$
\FOR{each $\mat{W}\in\{\mat{W}\}$}
    \STATE $\mat{W}=[\mat{W}_1\mid\cdots\mid \mat{W}_K]$, \quad $\mat{W}_k=\mat{L}_k\operatorname{diag}(\mat{s}_k)\mat{R}_k^\top$
\ENDFOR
\STATE $\bm{\theta}_0\gets\operatorname{concat}(\{\mat{s}_k\})$, \quad $\mat{D}\gets\operatorname{diag}(\mu_{\mathrm{rel}}|\bm{\theta}_0|)$
\STATE Freeze $\{\mat{L}_k,\mat{R}_k\}$
\FOR{$t=0,\ldots,T-1$}
    \FOR{$j=1,\ldots,N/2$}
        \STATE $\mat{z}_j\sim\mathcal{N}(0,\mat{I}_d)$
        \STATE $f_j^\pm\gets F(\bm{\theta}_t\pm \mat{D}\mat{z}_j)$
    \ENDFOR
    \STATE $\{\tilde f_j^\pm\}\gets \operatorname{Standardize}(\{f_j^\pm\})$
    \STATE $\displaystyle
    \bm{\theta}_{t+1}\gets\bm{\theta}_t+
    \frac{\alpha}{N}\mat{D}
    \sum_{j=1}^{N/2}
    (\tilde f_j^+-\tilde f_j^-)\mat{z}_j$
\ENDFOR
\STATE \textbf{return} $\bm{\theta}_T$
\end{algorithmic}
\end{minipage}
\end{algorithm}

\paragraph{Implementation.}
Experiments run on NVIDIA A100 40GB GPUs. For generation-intensive tasks, we use vLLM \cite{kwon2023vllm} to evaluate perturbed models. Population members are evaluated independently and in parallel. FuRA-S keeps an FP32 master copy of its block singular values and reconstructs perturbed dense weights for generation, so generation uses standard inference kernels. Supervised fine-tuning evaluates perturbations directly in the factored representation.

\begin{table}[H]
\caption{FuRA-S settings across experiments. Pairs denotes the number of antithetic perturbation pairs per update, Batch denotes the number of examples, problems, or puzzles evaluated per population member, and $\mat{L}_k$ indicates the storage precision of the frozen left singular factors. All ES experiments use $\mu_{\mathrm{rel}}=3\times10^{-2}$, with $\alpha=1$ except for per-user LaMP adaptation.}
\label{tab:hparams}
\centering
\small
\setlength{\tabcolsep}{4pt}
\begin{tabular}{lllcccrrl}
\toprule
Experiment & Model & Update & $\mu_{\mathrm{rel}}$ & Step & Pairs & Batch & Updates & $\mat{L}_k$ \\
\midrule
Countdown & Qwen2.5 0.5B--7B, Llama-3.2-1B & ES & $3\times10^{-2}$ & $\alpha=1$ & 15 & 200 & 500 & FP32 \\
MATH & Qwen2.5-Math-7B & ES & $3\times10^{-2}$ & $\alpha=1$ & 15 & 1,024 & 200 & FP32 \\
Countdown (INT4) & Qwen2.5 1.5B, 3B & ES & $3\times10^{-2}$ & $\alpha=1$ & 25 & 200 & 300 & FP32 \\
SFT & Llama-3-8B & ZO-Adam & $10^{-2}$ & $\eta=0.5$ & 8 & 16 & 5,987 & BF16/NF4 \\
Sudoku & Qwen3.5-4B & ES & $3\times10^{-2}$ & $\alpha=1$ & 16 & 32 & 100 & FP32 \\
LaMP (U) & Llama-3.1-8B, Qwen3.5-4B & ES & $3\times10^{-2}$ & $\alpha=1$ & 15 & 128 & 200 & BF16 \\
LaMP (T) & Llama-3.1-8B, Qwen3.5-4B & ES & $3\times10^{-2}$ & $\alpha=3$ & 15 & $\le64$ & $\le20$ & BF16 \\
\bottomrule
\end{tabular}
\end{table}

\subsection{RLVR on Countdown}
\label{app:countdown-setup}

\paragraph{Experimental protocol.}
We follow the Countdown benchmark of ES at Scale \cite{qiu2025es}, which contains 2,200 arithmetic reasoning problems. The first 200 problems are used for training and the remaining 2,000 for evaluation. At each iteration, all population members are evaluated on the same 200 training problems.

The prompt follows the plain-text format of ES at Scale without a chat template, requesting reasoning within \texttt{<think>} tags and a final expression within \texttt{<answer>} tags. The reward combines answer correctness and format compliance:
\begin{equation}
r=r_{\mathrm{answer}}+0.1r_{\mathrm{format}}.
\end{equation}
The answer reward equals one when the final expression uses exactly the given numbers and evaluates to the target, and zero otherwise. The format reward equals one for the required reasoning-and-answer structure. Otherwise, a reasoning block receives 0.1 and an answer block receives 0.5.

Training and evaluation use greedy decoding with at most 1,024 generated tokens. Each population member receives the mean reward across the 200 training problems. Test accuracy is the fraction of problems with a correct final answer.

\paragraph{FuRA-S and full-parameter ES.}
FuRA-S trains 245,760 block singular values on Qwen2.5-0.5B, 327,680 on Llama-3.2-1B, 508,928 on Qwen2.5-1.5B, 838,656 on Qwen2.5-3B, and 1,132,544 on Qwen2.5-7B. All five models use 15 antithetic pairs per iteration, $\mu_{\mathrm{rel}}=3\times10^{-2}$, and $\alpha=1$ for 500 iterations. We select these hyperparameters on Qwen2.5-1.5B and apply them unchanged to the other models.

The frozen spectral factors are stored in FP32, and the reconstructed weights are evaluated in BF16. For full-parameter ES, we reproduce ES at Scale using 30 independent Gaussian perturbations per iteration, with $\sigma=10^{-3}$ and $\alpha=5\times10^{-4}$. Full-parameter ES maintains an FP32 master copy of the model weights. Both methods evaluate 30 perturbed models per iteration, so each run generates 3,000,000 training responses.

\paragraph{LoRA-ES and VeRA-ES comparison.}
To compare the optimization spaces under the same ES procedure, we additionally optimize rank-4 LoRA adapters on the seven linear projections of each decoder layer. The adapters contain between 2.2M and 10.1M trainable parameters across the evaluated models, with the LoRA $B$ factors initialized to zero.

LoRA-ES uses 15 antithetic pairs per iteration, $\mu_{\mathrm{rel}}=3\times10^{-2}$, and $\alpha=1$, with the perturbation magnitude matched to FuRA-S for each weight matrix. Both methods use the same evaluation budget. The Qwen2.5-1.5B comparison is averaged over two runs, with seeds 42 and 1. The remaining model configurations use seed 42.

VeRA-ES applies VeRA to the same projections. Frozen random matrices of rank 1024 are shared across all layers, and each projection trains a scaling vector $\mat{d}$ initialized to 0.1 and a scaling vector $\mat{b}$ initialized to zero, which gives between 0.48M and 1.59M trainable parameters across the evaluated models. VeRA-ES uses the settings of LoRA-ES, with the perturbation of $\mat{b}$ matched to FuRA-S for each weight matrix and that of $\mat{d}$ set to $\mu_{\mathrm{rel}}$ times its initial value, and every model uses seed 42.

\subsection{RLVR on MATH}
\label{app:math-setup}

\begin{table}[H]
\caption{Greedy pass@1 (\%) of Qwen2.5-Math-7B on five math benchmarks under three checkpoint selections, with FuRA-S minus Full ES in parentheses. Avg-5 averages all five benchmarks, and Avg-3 averages MATH500, Minerva, and OlympiadBench.}
\label{tab:math-full}
\centering
\small
\begin{tabular}{llrrrrrrrr}
\toprule
Selection & Method & Iter. & AIME24 & AMC & MATH500 & Minerva & Olympiad & Avg-5 & Avg-3 \\
\midrule
\multirow{3}{*}{Final} & Full ES & 200 & 26.7 & 57.8 & 77.6 & 26.8 & 37.0 & \textbf{45.2} & 47.2 \\
 & LoRA-ES & 200 & 16.7 & 53.0 & 74.0 & 22.8 & 35.4 & 40.4 & 44.1 \\
 & \cellcolor{oursbg}FuRA-S & \cellcolor{oursbg}200 & \cellcolor{oursbg}23.3 & \cellcolor{oursbg}54.2 & \cellcolor{oursbg}76.8 & \cellcolor{oursbg}26.5 & \cellcolor{oursbg}40.0 & \cellcolor{oursbg}44.2 \negd{$-$1.0} & \cellcolor{oursbg}\textbf{47.8} \posd{+0.6} \\
\midrule
\multirow{3}{*}{Best Avg-5} & Full ES & 130 & 33.3 & 57.8 & 76.8 & 28.7 & 39.7 & \textbf{47.3} & 48.4 \\
 & LoRA-ES & 120 & 30.0 & 53.0 & 73.2 & 23.2 & 35.3 & 42.9 & 43.9 \\
 & \cellcolor{oursbg}FuRA-S & \cellcolor{oursbg}185 & \cellcolor{oursbg}23.3 & \cellcolor{oursbg}56.6 & \cellcolor{oursbg}77.2 & \cellcolor{oursbg}29.4 & \cellcolor{oursbg}40.1 & \cellcolor{oursbg}45.3 \negd{$-$1.9} & \cellcolor{oursbg}\textbf{48.9} \posd{+0.5} \\
\midrule
\multirow{3}{*}{Best Avg-3} & Full ES & 170 & 16.7 & 59.0 & 78.2 & 30.5 & 39.0 & 44.7 & \textbf{49.2} \\
 & LoRA-ES & 190 & 13.3 & 51.8 & 74.6 & 22.8 & 36.9 & 39.9 & 44.8 \\
 & \cellcolor{oursbg}FuRA-S & \cellcolor{oursbg}185 & \cellcolor{oursbg}23.3 & \cellcolor{oursbg}56.6 & \cellcolor{oursbg}77.2 & \cellcolor{oursbg}29.4 & \cellcolor{oursbg}40.1 & \cellcolor{oursbg}\textbf{45.3} \posd{+0.7} & \cellcolor{oursbg}48.9 \negd{$-$0.3} \\
\bottomrule
\end{tabular}
\end{table}

\paragraph{Experimental protocol.}
We follow the MATH training protocol of ES at Scale \cite{qiu2025es}, using 8,523 problems of difficulty levels 3--5 with Qwen2.5-Math-7B. Prompts use the model's chat template and instruct it to reason step by step and place the final answer in $\boxed{\phantom{x}}$. A response receives a binary reward of one if its final boxed answer is correct under the verifier of ES at Scale, and zero otherwise.

Each iteration evaluates 30 population members on the same 1,024 training prompts, selected sequentially from a fixed random permutation of the training set. Generation uses greedy decoding with a maximum of 3,000 new tokens.

We evaluate greedy pass@1 on AIME24 (30 problems), AMC (83), MATH500 (500), Minerva (272), and OlympiadBench (675), using the same answer verifier and generation limit. We evaluate at initialization and every five iterations. Table~\ref{tab:math-full} reports the final checkpoint at iteration 200 and the best checkpoints selected according to the five-benchmark average (Avg-5) or the average over MATH500, Minerva, and OlympiadBench (Avg-3). Avg-3 covers 1,447 problems and leaves out the two smallest sets, AIME24 and AMC.

\paragraph{Additional output-side coordinates.}
For the MATH experiment, FuRA-S adds a small set of trainable LM-head offsets to its block-spectral coordinates. These coordinates are constructed from the model's pretrained output weights and the dominant directions of the hidden states associated with answer generation.

We first generate responses to 512 randomly selected training problems using the pretrained model. From the resulting answers, we identify the 2,048 most frequent answer tokens, accounting for 98.5\% of observed answer-token occurrences. We also compute the leading 64 eigenvectors of the hidden-state second-moment matrix
\begin{equation}
    \sum_h \mat{h}\mat{h}^\top,
\end{equation}
where $\mat{h}\in\mathbb{R}^{d_h}$ denotes a hidden state entering the language-model head. These directions account for 83\% of the measured hidden-state energy.

Let $\mat{V}\in\mathbb{R}^{64\times d_h}$ contain the selected eigenvectors and let $\mat{H}\in\mathbb{R}^{2048\times d_h}$ denote the LM-head weight rows corresponding to the selected answer tokens. We introduce a trainable matrix $\mat{C}\in\mathbb{R}^{2048\times64}$, initialized to zero, and parameterize the selected output weights as
\begin{equation}
    \mat{H}'=\mat{H}+\mat{C}\mat{V}.
\end{equation}
The ES procedure jointly optimizes $\mat{C}$ and the block singular values. Perturbations of $\mat{C}$ are scaled elementwise by $\mu_{\mathrm{rel}}|\mat{H}\mat{V}^\top|$, following the relative scaling of the block singular values.

The resulting trainable state contains 131,072 output-side coordinates and 1,132,544 block singular values, for a total of 1,263,616 parameters.

\paragraph{Optimization and computational cost.}
FuRA-S, LoRA-ES, and full-parameter ES are trained for 200 iterations using the same sequence of training prompts. FuRA-S uses 15 antithetic pairs, $\mu_{\mathrm{rel}}=3\times10^{-2}$, and $\alpha=1$, with seed 42. Its spectral factors are stored in FP32 and reconstructed into BF16 weights for generation with vLLM.

Our full-parameter ES reproduction uses 30 independent perturbations with $\sigma=10^{-3}$ and $\alpha=5\times10^{-4}$, also with seed 42. It maintains an FP32 master copy of the model weights and evaluates perturbed models in BF16. LoRA-ES uses the rank-4 adapter configuration and perturbation procedure described in Appendix~\ref{app:countdown-setup}, with 10.1M trainable parameters.

On six NVIDIA A100 40GB GPUs, FuRA-S requires 732 seconds per iteration, compared with 739 seconds for full-parameter ES, as generation dominates both. The trainable state of FuRA-S occupies 5.1 MB, whereas full-parameter ES maintains a 30.5 GB FP32 master copy.

\subsection{RLVR on Quantized Models}
\label{app:quant-setup}

\paragraph{Experimental protocol.}
We follow the released Countdown protocol of QES \cite{xu2026qes}, using the same 200 training problems and 2,000 test problems as the full-precision experiments (Appendix~\ref{app:countdown-setup}). Both methods use identical Qwen2.5 checkpoints, prompt templates, reward definitions, and greedy decoding with a maximum of 1,024 generated tokens. The reward combines answer correctness with a weighted format reward, following the QES implementation.

Each iteration evaluates 50 population members on all 200 training problems. The unperturbed model is evaluated on the 2,000 test problems every 25 iterations, and results are reported after 300 training iterations. Evaluating the released INT4 checkpoints before adaptation with vLLM's GPTQ kernels yields initial accuracies of 3.45\% and 2.65\% for Qwen2.5-1.5B and Qwen2.5-3B, against the 3.50\% and 2.80\% reported by QES.

\paragraph{FuRA-S on released INT4 checkpoints.}
We adapt the released GPTQ-INT4 checkpoints of Qwen2.5-1.5B-Instruct and Qwen2.5-3B-Instruct, using the same quantized models as QES. Decoder projections use groupwise 4-bit quantization with FP16 scales and a group size of 128 input channels. Embeddings, normalization layers, and biases remain in FP16.

FuRA-S constructs its block-spectral coordinates from the dequantized weights, using the partitioning rule of Section~\ref{sec:coords} with block widths ranging from 48 to 128. The decomposition reconstructs the dequantized weights with relative error below $4\times10^{-5}$. The original INT4 weights and quantization parameters remain unchanged. The frozen $\mat{R}$ is stored in 4 bits, by symmetric round-to-nearest with one FP16 scale per column of each block.

The two models contain 508,928 and 838,656 trainable singular values, respectively. FuRA-S uses 25 antithetic pairs per iteration, which match the 50 members of QES, and the $\mu_{\mathrm{rel}}=3\times10^{-2}$ and $\alpha=1$ of the other reward-based experiments.

During training, perturbed weights are reconstructed from their spectral factors and evaluated in BF16 using vLLM. Section~\ref{sec:quant} and Table~\ref{tab:quant-curve} report accuracy under this evaluation. The INT4 deployment formulation of Section~\ref{sec:quant-method} instead applies the learned adaptation while retaining the released quantized weights and GPTQ kernels. Its memory and inference-time costs are reported in Appendix~\ref{app:cost}. On the first 200 test problems, the deployed checkpoints at iteration 300 solve 23.5\% and 46.0\%, against 22.5\% and 46.0\% in the BF16 evaluation, with the same outcome on 198 and 188 problems.

\begin{figure}[H]
\centering
\includegraphics[width=0.55\textwidth]{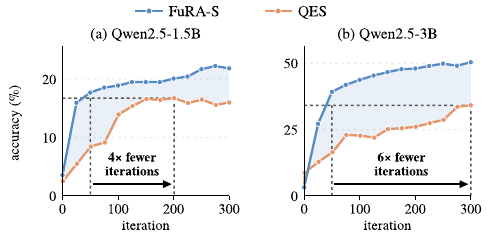}
\caption{Countdown accuracy of the INT4 checkpoints over iterations, with the best accuracy of QES dotted.}
\label{fig:quant}
\end{figure}

\paragraph{QES baseline.}
We reproduce QES using its released implementation and original optimization settings. Full-parameter QES is evaluated on two A100 40GB GPUs for Qwen2.5-1.5B and six for Qwen2.5-3B.

At iteration 300, the reproduced QES checkpoints achieve 16.00\% and 34.15\% accuracy, compared with 16.00\% and 31.85\% in the original QES report. QES reaches its best observed accuracy of 16.75\% after 201 updates on Qwen2.5-1.5B and 34.15\% after 300 updates on Qwen2.5-3B.

FuRA-S surpasses both values by iteration 50, reaching 17.70\% and 39.20\%, respectively (Table~\ref{tab:quant-curve}). QES evaluates the model after 1, 26, \dots, 276, and 300 updates, and Figure~\ref{fig:quant} plots its points at these update counts.

\begin{table}[H]
\caption{Countdown accuracy (\%) of FuRA-S on the GPTQ-INT4 checkpoints over training iterations, evaluated using reconstructed BF16 weights under the QES protocol. FuRA-S exceeds the best accuracy of our QES reproduction, 16.75\% and 34.15\%, within 50 iterations.}
\label{tab:quant-curve}
\centering
\small
\begin{tabular}{lrrrrrrrr}
\toprule
Iteration & 0 & 25 & 50 & 100 & 150 & 200 & 250 & 300 \\
\midrule
Qwen2.5-1.5B & 3.55 & 15.95 & 17.70 & 18.90 & 19.50 & 20.10 & 21.70 & 21.85 \\
Qwen2.5-3B & 3.25 & 27.15 & 39.20 & 43.75 & 46.65 & 48.00 & 49.85 & 50.40 \\
\bottomrule
\end{tabular}
\end{table}

\subsection{Mobile GUI Agents}
\label{app:gui}

\paragraph{Environment and model.} The task runs in a MobileWorld Android emulator in Docker \cite{kong2025mobileworld}, reset before every episode. The agent is MAI-UI-2B with its released system prompt, which asks for a thought in \texttt{<thinking>} tags and one \texttt{mobile\_use} tool call per step, with click, long press, type, swipe, open, drag, system button, wait, terminate, answer, and ask-user actions. Each step sees the task, the action history, and the recent screenshots. An episode ends on \texttt{terminate} or \texttt{answer}, or after 30 steps, and the task's checker scores the final device state with a partial credit between 0 and 1. An episode succeeds when its credit is 1. Generation runs in vLLM.

\paragraph{Step-wise fitness.} After each episode, every step is reduced to a screenshot hash and an action bucket. The hash is the 256-bit difference hash of the screenshot the agent acted on (grayscale, resized to $17\times16$, signs of the horizontal gradient). The bucket is the action type with the $108$-pixel grid cell for click-like actions, the direction for swipes, and the lower-cased argument for text, open-app, and key actions. Anchor groups are formed greedily over the pooled steps of the iteration: a step joins the first group whose representative lies within a Hamming distance of 10, and steps in a group visited by fewer than two distinct episodes receive zero advantage. We use $\gamma=0.95$. An iteration pools about 1,260 steps into 136 groups on average, 76 of them shared by at least two episodes, and 52\% of steps receive a nonzero advantage.

\paragraph{Optimization.} FuRA-S trains the block singular values of every linear projection of the language model, 516,096 values, with 8 antithetic pairs ($N=16$), $J=4$ greedy episodes per member, fitness standardized over the population, and $\alpha=1$, for 15 iterations. Greedy members still produce different episodes, because the emulator's loading times split their trajectories: in the first iteration, members succeed on 0 to 3 of their 4 episodes. Two vLLM servers share one GPU and drive 12 emulators. Malformed tool calls stay at or below 1.3\% of steps. Each task trains on its own GPU with the same settings, and an iteration takes 15 to 51 minutes, almost all of it rollouts. Episode-reward ES uses the same settings, with members sampled at temperature 0.7 under common random numbers per episode index, and scores each member by its mean partial credit over its 4 episodes.

\paragraph{Evaluation.} Each trained model is rebuilt from its perturbation seeds and update coefficients and runs 64 new episodes. Base and trained models are evaluated back to back on the same 12 emulators. The trained model completes MastodonReply in 11.5 steps on average. During training, the current model is also evaluated every 5 iterations on 16 episodes.

\subsection{Personalization on LaMP}
\label{app:lamp-setup}

\paragraph{Experimental protocol.}
We evaluate personalization on six LaMP tasks \cite{salemi2024lamp} under two splits. The user-based split (U) follows the official LaMP partition. FuRA-S trains on the training users, selects its checkpoint on the development users, and is tested on 2,500, 870, 2,500, 2,376, 2,500, and 1,496 queries for LaMP-1, 2, 3, 4, 5, and 7, respectively. Prompts follow the LaMP templates and include the top BM25 item from the user's own profile within a budget of 512 tokens, whereas the model without the profile receives the input alone.

The time-based split (T) follows the release of OPPU \cite{tan2024oppu}. For each task, we take the 100 users with the longest histories (101 for LaMP-3 and LaMP-5), which yields 123, 3,302, 112, 6,725, 107, and 109 test queries for LaMP-1, 2M, 3, 4, 5, and 7, respectively. Prompts follow the OPPU templates, with each field truncated to 768 words. Both splits apply the chat template of each model, with thinking disabled for Qwen3.5-4B, and limit sequences to 2,048 tokens.

All methods sample one answer per query with temperature 0.1, top-$k$ 10, and top-$p$ 0.9, generating at most 200 new tokens. We report accuracy and macro-F1 for LaMP-1, 2, and 2M, mean absolute error (MAE) and root mean squared error (RMSE) for LaMP-3, and ROUGE-1 and ROUGE-L for LaMP-4, 5, and 7.

\paragraph{Optimization settings.}
FuRA-S uses 15 antithetic pairs and $\mu_{\mathrm{rel}}=3\times10^{-2}$ in both splits, with the mean log-likelihood of the target tokens as fitness. In the user-based split, each task trains one model for 200 iterations with $\alpha=1$, and each iteration evaluates 128 random training examples. The unperturbed model is scored on 256 fixed development examples at initialization and every 20 iterations, and the best state is retained.

In the time-based split, each user's checkpoint starts from the base model and runs for at most 20 iterations with $\alpha=3$, each on up to 64 examples from the user's history. The latest 10\% of the history, with at least 8 records and at most half of the history, serves as validation. Training stops after three iterations without improvement, and the best state answers the test queries. The frozen $\mat{L}_k$ factors are stored in BF16, while $\mat{s}_k$ and $\mat{R}_k$ are maintained in FP32.

\paragraph{Per-user LoRA baseline.}
We reimplement the per-user stage of OPPU. Each user receives a new rank-8 LoRA adapter on the query and value projections, trained from the base model for two epochs over the user's history. Training uses AdamW with a learning rate of $10^{-4}$, weight decay of 0.01, batch size 16, LoRA scale 8, dropout 0.05, 10\% warm-up followed by linear decay, and gradient clipping at 0.3.

\paragraph{Computational cost.}
Each setting runs once with seed 42 on nodes with four NVIDIA A100 40GB GPUs. On Llama-3.1-8B-Instruct, a user-based model takes approximately 0.9 node-hours per task and up to 2.0 node-hours for LaMP-3.

\section{System Cost}
\label{app:cost}

\paragraph{Measurement protocol.}
We measure training memory and runtime on Llama-3-8B under the Commonsense-170K setting (Appendix~\ref{app:sft}), using an NVIDIA A100 40GB GPU. Peak memory is measured over training steps using the same 128 examples and 16-example mini-batches for all methods. Runtime includes perturbation, forward evaluation, and parameter updates, measured over 384 training examples. We evaluate both length-grouped batches and conventionally padded batches to distinguish optimization costs from sequence-padding overhead. GRZO evaluates its loss per example and therefore runs only with padded batches.

\begin{table}[H]
\caption{Peak training memory and time per 16-example forward evaluation on Llama-3-8B using one A100 40GB GPU. Runtime includes perturbation and parameter updates, with length-grouped or padded batches.}
\label{tab:cost}
\centering
\small
\begin{tabular}{lrrr}
\toprule
Method & Peak memory (GB) & Grouped (s) & Padded (s) \\
\midrule
FuRA-S & 17.17 & 0.341 & 0.570 \\
FuRA-S (4-bit $\mat{L}$) & 7.52 & 0.358 & 0.579 \\
MeZO & 16.26 & 0.318 & 0.481 \\
LOZO & 16.26 & 0.265 & 0.427 \\
GRZO & 20.90 & -- & 1.028 \\
Sparse-MeZO & 30.90 & 0.408 & 0.570 \\
VeRA & 16.93 & 0.379 & 0.656 \\
Uni-LoRA & 16.49 & 0.261 & 0.436 \\
QuZO (4-bit) & 6.71 & 0.325 & 0.493 \\
\bottomrule
\end{tabular}
\end{table}

\paragraph{Training memory.}
Forward-only training stores no activations for backpropagation. FuRA-S further limits its trainable state to the block singular values and their optimizer states, requiring approximately 20 MB beyond forward evaluation. With BF16 singular factors, peak training memory reaches 17.17 GB, compared with 17.15 GB for forward evaluation alone. Quantizing the frozen left singular factors to NF4 reduces training memory to 7.52 GB, approximately 46\% of the memory required by full-parameter MeZO (Table~\ref{tab:cost}).

\paragraph{Training time.}
FuRA-S has forward-evaluation costs comparable to other zeroth-order methods. On length-grouped mini-batches, a forward evaluation including perturbation and update takes 0.341 seconds, increasing to 0.358 seconds with NF4 left singular factors. On MATH, an iteration takes 732 seconds for FuRA-S and 739 seconds for full-parameter ES on six A100 40GB GPUs, because generation dominates its cost.

\paragraph{Adaptation storage.}
For full-precision models, the adapted blocks can be merged into dense weights without introducing additional inference operations. Alternatively, multiple adaptations can share the same frozen factors and store only their optimized block singular values. A FuRA-S checkpoint for Llama-3-8B occupies 4.98 MB in FP32, compared with 16.1 GB for a BF16 copy of the entire model. For the INT4 checkpoints of Section~\ref{sec:quant}, the released 4-bit weights remain unchanged, and the 4-bit $\mat{R}$ and FP32 block singular values raise the storage of the projections from 4.16 to 4.29--4.31 bits per weight.

\paragraph{INT4 inference overhead.}
We evaluate the released GPTQ-INT4 Qwen2.5-1.5B and Qwen2.5-3B checkpoints from Section~\ref{sec:quant} on one NVIDIA A100 40GB GPU. All configurations use the same GPTQ kernels in vLLM \cite{kwon2023vllm}, with 4-bit weights and FP16 activations. Decoding is measured over 256 generated tokens following a 128-token prompt at batch size one, using a key-value cache and captured CUDA graphs. Prefill is measured using one or eight prompts of 512 tokens.

We compare the original INT4 models, the deployed FuRA-S formulation, and two training-time representations with FP16 or FP32 right singular factors. The deployed formulation folds the learned blockwise correction into a single batched operation preceding each original INT4 weight multiplication. During training, the right singular factors and block singular values remain separate, requiring two batched operations.

Relative to the INT4 baseline, the deployed formulation increases GPU memory by approximately 9\% and decoding latency by 11--15\%, while retaining a memory footprint 2.4--2.8 times smaller than FP16. Prefill incurs a larger overhead, approximately 50\% for a single prompt and 18--21\% for eight prompts (Table~\ref{tab:quant-deploy}).

\begin{table}[H]
\caption{INT4 inference cost on one A100 40GB GPU. Memory is reported in GB, decoding latency per token in milliseconds, and prefill latency in milliseconds for one or eight prompts. The deployed FuRA-S formulation retains the original GPTQ weights and introduces approximately 9\% additional memory relative to INT4.}
\label{tab:quant-deploy}
\centering
\small
\begin{tabular}{lrrrrrrrr}
\toprule
 & \multicolumn{4}{c}{Qwen2.5-1.5B} & \multicolumn{4}{c}{Qwen2.5-3B} \\
\cmidrule(lr){2-5}\cmidrule(lr){6-9}
Model form & Memory & Decode & Prefill 1 & Prefill 8 & Memory & Decode & Prefill 1 & Prefill 8 \\
\midrule
FP16 & 2.88 & 6.93 & 23.9 & 80.5 & 5.85 & 10.58 & 31.9 & 156.9 \\
INT4 & 1.09 & 5.74 & 31.2 & 87.4 & 1.94 & 7.95 & 42.2 & 160.8 \\
INT4 + FuRA-S, deployed & 1.19 & 6.58 & 46.6 & 105.6 & 2.11 & 8.83 & 63.5 & 189.1 \\
INT4 + FuRA-S, training, FP16 $\mat{R}_k$ & 1.20 & 7.94 & 56.7 & 125.2 & 2.12 & 10.34 & 74.4 & 220.0 \\
INT4 + FuRA-S, training, FP32 $\mat{R}_k$ & 1.30 & 8.92 & 62.8 & 209.0 & 2.28 & 11.85 & 81.7 & 355.7 \\
\bottomrule
\end{tabular}
\end{table}

\section{Supervised Fine-Tuning}
\label{app:sft}

For supervised fine-tuning on Commonsense-170K, FuRA-S keeps the perturbations of Section~\ref{sec:optim} and replaces the ES update with the two-point gradient estimate of MeZO over $P$ antithetic pairs, each evaluated on its own mini-batch $\mathcal{B}_j$,
\begin{equation}
\hat{\mat{g}}_t=\frac{1}{P}\sum_{j=1}^{P}\frac{\mathcal{L}(\bm{\theta}_t+\mat{D}\mat{z}_j;\mathcal{B}_j)-\mathcal{L}(\bm{\theta}_t-\mat{D}\mat{z}_j;\mathcal{B}_j)}{2}\,\mat{z}_j,
\label{eq:zo-grad}
\end{equation}
to which it applies Adam~\cite{kingma2015adam} before scaling the update by $\mat{D}$.

We compare it with full-parameter ZO methods, parameter-efficient baselines, and 4-bit QuZO \cite{zhou2025quzo} under the same budget of forward evaluations.

\paragraph{Experimental protocol.}
We fine-tune Llama-3-8B on Commonsense-170K from LLM-Adapters \cite{hu2023llmadapters}, following the prompt and answer format of FuRA \cite{zhao2026fura}. We reserve 120 training examples for validation and optimize the mean negative log-likelihood of response tokens, with a maximum sequence length of 2,048 tokens. All methods use the same sequence of 16-example mini-batches drawn from a fixed random permutation of the training set.

Evaluation follows FuRA, using beam search with four beams and a maximum of 256 generated tokens. The first label word in the generated response determines the predicted answer. We evaluate the final checkpoint on PIQA (1,838 examples), SIQA (1,954), HellaSwag (10,042), WinoGrande (1,267), ARC-e (2,376), ARC-c (1,172), and OBQA (500). Each method is evaluated in one run with seed 43 on a single A100 40GB GPU.

\paragraph{Training budget.}
We match the computational budget of first-order FuRA, which performs three training epochs with 31,932 updates on 16-example mini-batches. Approximating one forward-backward update as three forward evaluations yields a budget of 95,796 forward evaluations, equivalent to 47,898 antithetic perturbation pairs.

All ZO methods in Table~\ref{tab:sft} receive this budget. Baselines using one perturbation pair per update perform 47,898 updates, while FuRA-S evaluates eight pairs per update for 5,987 updates. In the full-precision setting, each FuRA-S pair uses a separate mini-batch, resulting in the same 4.5 effective data passes as the ZO baselines. In the 4-bit frozen-factor setting, all eight pairs within an update share the same mini-batch.

\paragraph{Optimization settings.}
FuRA-S uses a relative perturbation scale of $\mu_{\mathrm{rel}}=10^{-2}$ and ZO-Adam with $\beta_1=0.9$, $\beta_2=0.999$, and $\epsilon=10^{-8}$. The peak learning rate is 0.5, with linear warm-up over the first 3\% of updates followed by linear decay to zero.

For full-precision adaptation, the frozen $\mat{L}_k$ factors are stored in BF16, while $\mat{s}_k$ and $\mat{R}_k$ are maintained in FP32. In the 4-bit frozen-factor variant, $\mat{L}_k$ is stored using NF4 with double quantization and a block size of 64. The block singular values and right singular factors remain in FP32.

Table~\ref{tab:sft-hparams} provides the optimization settings for all baselines. VeRA and Uni-LoRA have no zeroth-order recipe of their own, so we train them with the ZO-SGD procedure of MeZO.

\begin{table}[H]
\caption{Optimization settings for Commonsense-170K. LR denotes the learning rate, with peak values for FuRA-S, the initial value for QuZO, and constant values otherwise. All methods use the same budget of 47,898 antithetic perturbation pairs.}
\label{tab:sft-hparams}
\centering
\small
\setlength{\tabcolsep}{4pt}
\begin{tabular}{lllll}
\toprule
Method & Update & LR & Perturbation & Other settings \\
\midrule
FuRA-S & ZO-Adam & 0.5 & $\mu_{\mathrm{rel}}=10^{-2}$ & per-pair batches, BF16 $\mat{L}_k$ \\
FuRA-S (4-bit $\mat{L}$) & ZO-Adam & 0.5 & $\mu_{\mathrm{rel}}=10^{-2}$ & shared batch, NF4 $\mat{L}_k$ \\
GRZO & GRZO & $5\times10^{-8}$ & $\epsilon=10^{-3}$ & per-example perturbations, FP16 \\
LOZO & ZO-SGD & $10^{-7}$ & $\epsilon=10^{-3}$ & rank 2, $\nu=50$, FP16 \\
MeZO & ZO-SGD & $10^{-7}$ & $\epsilon=10^{-3}$ & FP16 \\
VeRA & ZO-SGD & $5\times10^{-3}$ & $\epsilon=10^{-2}$ & $r=1024$, $d_0=0.1$, FP16 \\
Uni-LoRA & ZO-SGD & $10^{-4}$ & $\epsilon=10^{-2}$ & $d=524{,}288$, LoRA rank 4, FP16 \\
Sparse-MeZO & ZO-SGD & $10^{-6}$ & $\epsilon=10^{-3}$ & sparsity 0.9, FP16 \\
QuZO & ZO-SGD & $5\times10^{-5}$ & $\epsilon=2\times10^{-3}$ & LoRA rank 8 on $q,v$, NF4 base \\
\bottomrule
\end{tabular}
\end{table}

\paragraph{Results.}
FuRA-S achieves 64.64\% average accuracy while training only 1.25M parameters (0.0155\% of the model), outperforming all parameter-efficient baselines (Table~\ref{tab:sft}). Across the seven benchmarks, it ranks second only to GRZO on six tasks and achieves the highest accuracy on WinoGrande. It exceeds VeRA by 2.98 points and full-parameter MeZO by 9.93 points. With its frozen left singular factors stored in 4 bits, FuRA-S retains 59.97\% accuracy, outperforming QuZO by 16.97 points.

\begin{table}[H]
\caption{Accuracy (\%) on seven tasks after SFT of Llama-3-8B on Commonsense-170K with the same forward-pass budget, best per group in bold. FuRA-S is the most accurate PEFT method and leads QuZO by 16.97 points in 4 bits.}
\label{tab:sft}
\centering
\resizebox{\textwidth}{!}{%
\begin{tabular}{lrrrrrrrrrr}
\toprule
Method & Trainable & Trainable \% & PIQA & SIQA & HellaS. & WinoG. & ARC-e & ARC-c & OBQA & Avg \\
\midrule
\midrule
\multicolumn{11}{l}{\emph{Full-parameter updates}} \\
GRZO & 8.03B & 100 & \textbf{78.7} & \textbf{64.4} & \textbf{74.3} & \textbf{53.0} & \textbf{84.2} & \textbf{65.7} & \textbf{60.6} & \textbf{68.70} \\
LOZO & 8.03B & 100 & 70.8 & 53.2 & 45.0 & 49.3 & 68.8 & 53.0 & 52.0 & 56.01 \\
MeZO & 8.03B & 100 & 76.2 & 55.2 & 40.5 & 51.5 & 67.0 & 51.8 & 40.8 & 54.71 \\
\midrule
\multicolumn{11}{l}{\emph{PEFT updates}} \\
\rowcolor{oursbg}
FuRA-S & 1.25M & 0.0155 & \textbf{78.0} & \textbf{56.2} & \textbf{66.3} & \textbf{53.4} & \textbf{79.7} & \textbf{61.3} & \textbf{57.6} & \textbf{64.64} \\
VeRA & 1.61M & 0.0200 & 76.3 & 54.0 & 60.1 & 50.7 & 77.2 & 60.7 & 52.6 & 61.66 \\
Uni-LoRA & 0.52M & 0.0065 & 76.3 & 52.1 & 45.1 & 52.2 & 74.8 & 56.1 & 54.8 & 58.77 \\
Sparse-MeZO & 0.80B & 10 & 74.0 & 47.4 & 38.6 & 51.2 & 71.0 & 57.8 & 53.8 & 56.26 \\
\midrule
\multicolumn{11}{l}{\emph{4-bit comparison}} \\
\rowcolor{oursbg}
FuRA-S (4-bit L) & 1.25M & 0.0155 & \textbf{73.7} & \textbf{54.0} & \textbf{53.8} & \textbf{52.9} & \textbf{72.5} & \textbf{58.5} & \textbf{54.4} & \textbf{59.97} \\
QuZO (4-bit) & 3.41M & 0.0424 & 53.1 & 45.3 & 31.7 & 50.1 & 48.9 & 40.3 & 31.6 & 43.00 \\
\bottomrule
\end{tabular}
}
\end{table}

\section{Long-Horizon Agentic Optimization}
\label{app:sudoku}

We evaluate FuRA-S on a long-horizon agentic task, where the agent acts over up to 45 turns and receives a single reward only when it completes the task.

\paragraph{Experimental protocol.}
We follow the multi-turn Sudoku benchmark of Agentic ESOpt \cite{zheng2026agenticesopt}, using its 15-cell completion setting. Each agent must fill 15 missing cells of a $9\times9$ board over at most 45 turns, with a maximum of 64 generated tokens per turn. A reward is provided only when the completed board is valid. The benchmark contains 32 training puzzles and 32 held-out puzzles.

We evaluate Qwen3.5-4B in BF16 with thinking disabled. The model has 32 decoder layers comprising attention and gated-DeltaNet components. FuRA-S adapts the linear projections in both components and the MLPs, covering 248 weight matrices. Each population member is scored by its success rate over the 32 training puzzles, and fitness is standardized across the population. Generation uses temperature 0.7, top-$p$ 0.8, top-$k$ 20, and presence penalty 1.5. Held-out performance is evaluated every 10 iterations with three repeated evaluations.

\paragraph{Optimization settings and comparison.}
FuRA-S trains 835,584 block singular values using 16 antithetic pairs per iteration, $\mu_{\mathrm{rel}}=3\times10^{-2}$, and $\alpha=1$. Training runs for 100 iterations. The full-parameter Vanilla ES baseline perturbs all 4.2B model parameters using 32 one-sided Gaussian perturbations per iteration, with $\sigma=\alpha=5\times10^{-4}$. Both settings use the same benchmark and number of population evaluations. Agentic ESOpt additionally decays the perturbation scale with a cosine schedule. FuRA-S and the Vanilla ES baseline both use a constant scale, so the comparison isolates the effect of the search space.

\begin{figure}[H]
\centering
\includegraphics[width=0.5\textwidth]{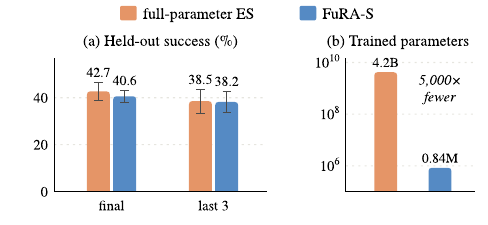}
\caption{Agentic Sudoku with Qwen3.5-4B: (a)~held-out success, final and over the last three evaluations, with standard deviations; (b)~trained parameters. FuRA-S is on par with full-parameter ES while training 5,000 times fewer parameters. Results for full-parameter ES are from the released logs of \citet{zheng2026agenticesopt}.}
\label{fig:sudoku}
\end{figure}

\paragraph{Results.}
FuRA-S matches the performance of full-parameter ES while optimizing only 835,584 block singular values, approximately 0.02\% of Qwen3.5-4B's parameters (Figure~\ref{fig:sudoku}). Both methods achieve approximately 38\% held-out success averaged over the final three evaluations. The trainable state of FuRA-S occupies 3.3 MB, against the 8.4 GB of weights that full-parameter ES updates.

\section{LaMP Scores with Llama-3.1-8B-Instruct}
\label{app:lamp-llama}

Tables~\ref{tab:lamp-u-llama} and~\ref{tab:lamp-t-llama} report the scores behind Figure~\ref{fig:lamp}, together with a secondary metric for each task. In the user-based split, FuRA-S with the retrieved profile improves over the base model with the same profile on all six tasks, with the largest gain on LaMP-5, where ROUGE-1 rises from 0.132 to 0.458. In the time-based split, the per-user FuRA-S checkpoints lead per-user LoRA on LaMP-2M and LaMP-7, whereas per-user LoRA leads on LaMP-1, LaMP-3, LaMP-4, and LaMP-5.

\begin{table}[H]
\caption{LaMP user-based split with Llama-3.1-8B-Instruct, with or without one retrieved item from the user's profile in the prompt, for FuRA-S in both training and testing. With the profile, FuRA-S improves over the base model on all six tasks.}
\label{tab:lamp-u-llama}
\centering
\footnotesize
\setlength{\tabcolsep}{4pt}
\resizebox{\textwidth}{!}{%
\begin{tabular}{llrrrr}
\toprule
Task & Metric & Base & Base + profile & FuRA-S & FuRA-S + profile \\
\midrule
LaMP-1 citation & accuracy / F1 & 0.528 / 0.528 & 0.665 / 0.666 & 0.515 / 0.515 & 0.681 / 0.681 \\
LaMP-2 movie tag & accuracy / F1 & 0.285 / 0.267 & 0.347 / 0.302 & 0.446 / 0.404 & 0.452 / 0.404 \\
LaMP-3 rating & MAE / RMSE $\downarrow$ & 0.383 / 0.697 & 0.369 / 0.748 & 0.272 / 0.577 & 0.254 / 0.564 \\
LaMP-4 news headline & ROUGE-1 / ROUGE-L & 0.093 / 0.081 & 0.123 / 0.109 & 0.165 / 0.150 & 0.182 / 0.166 \\
LaMP-5 paper title & ROUGE-1 / ROUGE-L & 0.118 / 0.109 & 0.132 / 0.120 & 0.438 / 0.398 & 0.458 / 0.418 \\
LaMP-7 tweet paraphrase & ROUGE-1 / ROUGE-L & 0.335 / 0.279 & 0.335 / 0.284 & 0.476 / 0.418 & 0.470 / 0.414 \\
\bottomrule
\end{tabular}
}
\end{table}

\begin{table}[H]
\caption{LaMP time-based split with Llama-3.1-8B-Instruct: the base model and per-user LoRA and FuRA-S checkpoints. FuRA-S leads per-user LoRA on LaMP-2M and LaMP-7.}
\label{tab:lamp-t-llama}
\centering
\footnotesize
\setlength{\tabcolsep}{4pt}
\resizebox{\textwidth}{!}{%
\begin{tabular}{llrrrr}
\toprule
Task & Metric & Queries (users) & Base & Per-user LoRA & FuRA-S \\
\midrule
LaMP-1 citation & accuracy / F1 & 123 (100) & 0.480 / 0.478 & 0.488 / 0.486 & 0.480 / 0.470 \\
LaMP-2M movie tag & accuracy / F1 & 3302 (100) & 0.296 / 0.256 & 0.491 / 0.357 & 0.541 / 0.422 \\
LaMP-3 rating & MAE / RMSE $\downarrow$ & 112 (101) & 0.402 / 0.840 & 0.179 / 0.463 & 0.259 / 0.701 \\
LaMP-4 news headline & ROUGE-1 / ROUGE-L & 6725 (100) & 0.090 / 0.078 & 0.164 / 0.147 & 0.141 / 0.125 \\
LaMP-5 paper title & ROUGE-1 / ROUGE-L & 107 (101) & 0.145 / 0.131 & 0.493 / 0.445 & 0.476 / 0.415 \\
LaMP-7 tweet paraphrase & ROUGE-1 / ROUGE-L & 109 (100) & 0.296 / 0.246 & 0.320 / 0.266 & 0.338 / 0.290 \\
\bottomrule
\end{tabular}
}
\end{table}

\section{LaMP Scores with Qwen3.5-4B}
\label{app:lamp-qwen}

Qwen3.5-4B adds gated-DeltaNet layers to attention layers. The per-user LoRA baseline adapts the query and value projections of attention layers and the fused query-key-value projections of gated-DeltaNet layers.

\begin{figure}[H]
\centering
\includegraphics[width=\textwidth]{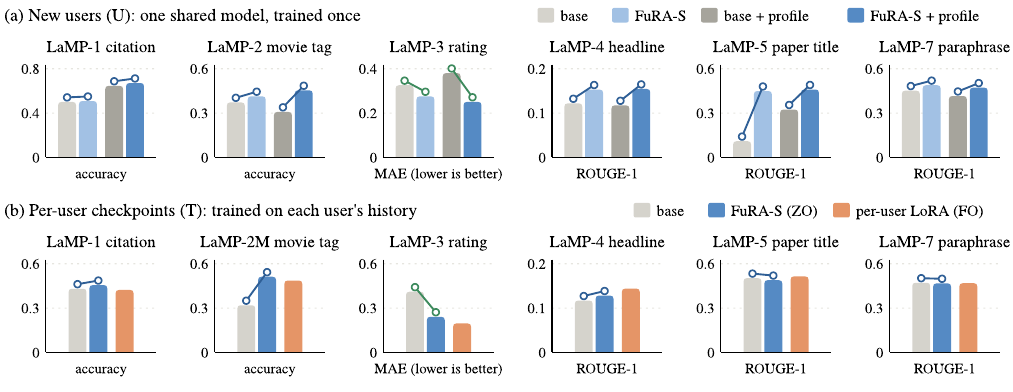}
\caption{LaMP with Qwen3.5-4B, with the panels of Figure~\ref{fig:lamp}. With the profile, FuRA-S again improves all six tasks for new users, and its per-user checkpoints lead per-user LoRA on LaMP-1 and LaMP-2M.}
\label{fig:lamp-qwen}
\end{figure}

In the user-based split, FuRA-S with the retrieved profile improves over the base model with the same profile on all six tasks (Figure~\ref{fig:lamp-qwen} and Table~\ref{tab:lamp-u-qwen}). In the time-based split, FuRA-S leads per-user LoRA on LaMP-1 and LaMP-2M, and per-user LoRA leads on the other four tasks (Table~\ref{tab:lamp-t-qwen}). On LaMP-7, the base model, per-user LoRA, and FuRA-S score within 0.004 ROUGE-1 of one another.

\begin{table}[H]
\caption{LaMP user-based split with Qwen3.5-4B, with or without one retrieved item from the user's profile in the prompt, for FuRA-S in both training and testing. With the profile, FuRA-S improves over the base model on all six tasks.}
\label{tab:lamp-u-qwen}
\centering
\footnotesize
\setlength{\tabcolsep}{4pt}
\resizebox{\textwidth}{!}{%
\begin{tabular}{llrrrr}
\toprule
Task & Metric & Base & Base + profile & FuRA-S & FuRA-S + profile \\
\midrule
LaMP-1 citation & accuracy / F1 & 0.500 / 0.511 & 0.646 / 0.642 & 0.509 / 0.498 & 0.671 / 0.670 \\
LaMP-2 movie tag & accuracy / F1 & 0.372 / 0.304 & 0.308 / 0.257 & 0.414 / 0.350 & 0.454 / 0.388 \\
LaMP-3 rating & MAE / RMSE $\downarrow$ & 0.326 / 0.675 & 0.381 / 0.826 & 0.276 / 0.607 & 0.251 / 0.578 \\
LaMP-4 news headline & ROUGE-1 / ROUGE-L & 0.122 / 0.107 & 0.117 / 0.103 & 0.153 / 0.139 & 0.154 / 0.141 \\
LaMP-5 paper title & ROUGE-1 / ROUGE-L & 0.110 / 0.099 & 0.324 / 0.281 & 0.449 / 0.410 & 0.460 / 0.420 \\
LaMP-7 tweet paraphrase & ROUGE-1 / ROUGE-L & 0.452 / 0.396 & 0.415 / 0.363 & 0.489 / 0.432 & 0.472 / 0.419 \\
\bottomrule
\end{tabular}
}
\end{table}

\begin{table}[H]
\caption{LaMP time-based split with Qwen3.5-4B: the base model and per-user LoRA and FuRA-S checkpoints. FuRA-S leads per-user LoRA on LaMP-1 and LaMP-2M.}
\label{tab:lamp-t-qwen}
\centering
\footnotesize
\setlength{\tabcolsep}{4pt}
\resizebox{\textwidth}{!}{%
\begin{tabular}{llrrrr}
\toprule
Task & Metric & Queries (users) & Base & Per-user LoRA & FuRA-S \\
\midrule
LaMP-1 citation & accuracy / F1 & 123 (100) & 0.431 / 0.446 & 0.423 / 0.425 & 0.455 / 0.458 \\
LaMP-2M movie tag & accuracy / F1 & 3302 (100) & 0.319 / 0.248 & 0.486 / 0.334 & 0.512 / 0.385 \\
LaMP-3 rating & MAE / RMSE $\downarrow$ & 112 (101) & 0.411 / 0.732 & 0.196 / 0.567 & 0.241 / 0.620 \\
LaMP-4 news headline & ROUGE-1 / ROUGE-L & 6725 (100) & 0.117 / 0.104 & 0.144 / 0.129 & 0.128 / 0.115 \\
LaMP-5 paper title & ROUGE-1 / ROUGE-L & 107 (101) & 0.502 / 0.422 & 0.514 / 0.457 & 0.489 / 0.432 \\
LaMP-7 tweet paraphrase & ROUGE-1 / ROUGE-L & 109 (100) & 0.471 / 0.421 & 0.468 / 0.421 & 0.467 / 0.419 \\
\bottomrule
\end{tabular}
}
\end{table}

\end{document}